\documentclass[conference]{IEEEtran}
\IEEEoverridecommandlockouts
\usepackage{cite}
\usepackage{amsmath,amssymb,amsfonts}
\usepackage{algorithmic}
\RequirePackage{bm} 
\RequirePackage{url}
\usepackage{multirow}
\usepackage{subfigure}
\usepackage{makecell}
\usepackage{array}
\usepackage{algorithm}
\usepackage{algorithmic}
\usepackage{dsfont}
\usepackage{cleveref}
\usepackage{graphicx}
\usepackage{textcomp}
\usepackage{xcolor}
\usepackage{mathtools}

\usepackage{booktabs}

\ifx\BlackBox\undefined
\newcommand{\BlackBox}{\rule{1.5ex}{1.5ex}}  
\fi

\ifx\QED\undefined
\def\QED{~\rule[-1pt]{5pt}{5pt}\par\medskip}
\fi

\ifx\proof\undefined
\newenvironment{proof}{\par\noindent{\bf Proof\ }}{\hfill\BlackBox\\[2mm]}
\fi

\ifx\theorem\undefined
\newtheorem{theorem}{Theorem}
\fi
\ifx\example\undefined
\newtheorem{example}[theorem]{Example}
\fi
\ifx\property\undefined
\newtheorem{property}{Property}
\fi
\ifx\lemma\undefined
\newtheorem{lemma}[theorem]{Lemma}
\fi
\ifx\proposition\undefined
\newtheorem{proposition}[theorem]{Proposition}
\fi
\ifx\remark\undefined
\newtheorem{remark}[theorem]{Remark}
\fi
\ifx\corollary\undefined
\newtheorem{corollary}[theorem]{Corollary}
\fi
\ifx\definition\undefined
\newtheorem{definition}[theorem]{Definition}
\fi
\ifx\conjecture\undefined
\newtheorem{conjecture}[theorem]{Conjecture}
\fi
\ifx\fact\undefined
\newtheorem{fact}[theorem]{Fact}
\fi
\ifx\claim\undefined
\newtheorem{claim}[theorem]{Claim}
\fi
\ifx\assumption\undefined
\newtheorem{assumption}[theorem]{Assumption}
\fi
\ifx\condition\undefined
\newtheorem{condition}[theorem]{Condition}
\fi

\newcommand{\argmax}{\mathop{\mathrm{argmax}}}
\def\BibTeX{{\rm B\kern-.05em{\sc i\kern-.025em b}\kern-.08em
    T\kern-.1667em\lower.7ex\hbox{E}\kern-.125emX}}
\begin{document}

\title{$\epsilon$-Neural Thompson Sampling of Deep Brain Stimulation for Parkinson Disease Treatment\\
}

\author{\IEEEauthorblockN{ Hao-Lun Hsu}
\IEEEauthorblockA{\textit{Computer Science} \\
\textit{Duke University}\\
\textit{Durham, NC, USA}\\
hao-lun.hsu@duke.edu}
\and
\IEEEauthorblockN{Qitong Gao}
\IEEEauthorblockA{\textit{Electrical and Computer Engineering} \\
\textit{Duke University}\\
\textit{Durham, NC, USA}\\
qitong.gao@duke.edu}
\and
\IEEEauthorblockN{Miroslav Pajic}
\IEEEauthorblockA{\textit{Electrical and Computer Engineering} \\
\textit{Duke University}\\
\textit{Durham, NC, USA}\\
miroslav.pajic@duke.edu}
\thanks{This work is sponsored in part by the NSF CNS-1837499 award and the National AI Institute for Edge Computing Leveraging Next Generation Wireless Networks, Grant CNS-2112562, as well as by NIH UH3 NS103468.}
}

\maketitle

\begin{abstract}
Deep Brain Stimulation (DBS) stands as an effective intervention for alleviating the motor symptoms of Parkinson's disease (PD). Traditional commercial DBS devices are only able to deliver fixed-frequency periodic pulses to the basal ganglia (BG) regions of the brain, \textit{i.e.}, continuous DBS (cDBS). However, they in general suffer from energy inefficiency and side effects, such as speech impairment. Recent research has focused on adaptive DBS (aDBS) to resolve the limitations of cDBS. Specifically, reinforcement learning (RL) based approaches have been developed to adapt the frequencies of the stimuli in order to achieve both energy efficiency and treatment efficacy. However, RL approaches in general require significant amount of training data and computational resources, making it intractable to integrate RL policies into real-time embedded systems as needed in aDBS. In contrast, contextual multi-armed bandits (CMAB) in general lead to better sample efficiency compared to RL. In this study, we propose a CMAB solution for aDBS. Specifically, we define the context as the signals capturing irregular neuronal firing activities in the BG regions (\textit{i.e.}, beta-band power spectral density), while each `arm' signifies the (discretized) pulse frequency of the stimulation. Moreover, an $\epsilon$-exploring strategy is introduced on top of the classic Thompson sampling method, leading to an algorithm called $\epsilon$-Neural Thompson sampling ($\epsilon$-NeuralTS), such that the learned CMAB policy can better balance exploration and exploitation of the BG environment. The $\epsilon$-NeuralTS algorithm is evaluated using a computation BG model that captures the neuronal activities in PD patients' brains. The results show that our method outperforms both existing cDBS methods, as well as the baselines that do not use the $\epsilon$-exploring as introduced by our method (\textit{i.e.,} the vanilla Thompson sampling~method).
\end{abstract}

\section{Introduction}
\begin{figure}
    \centering
    \includegraphics[width = .42\textwidth]{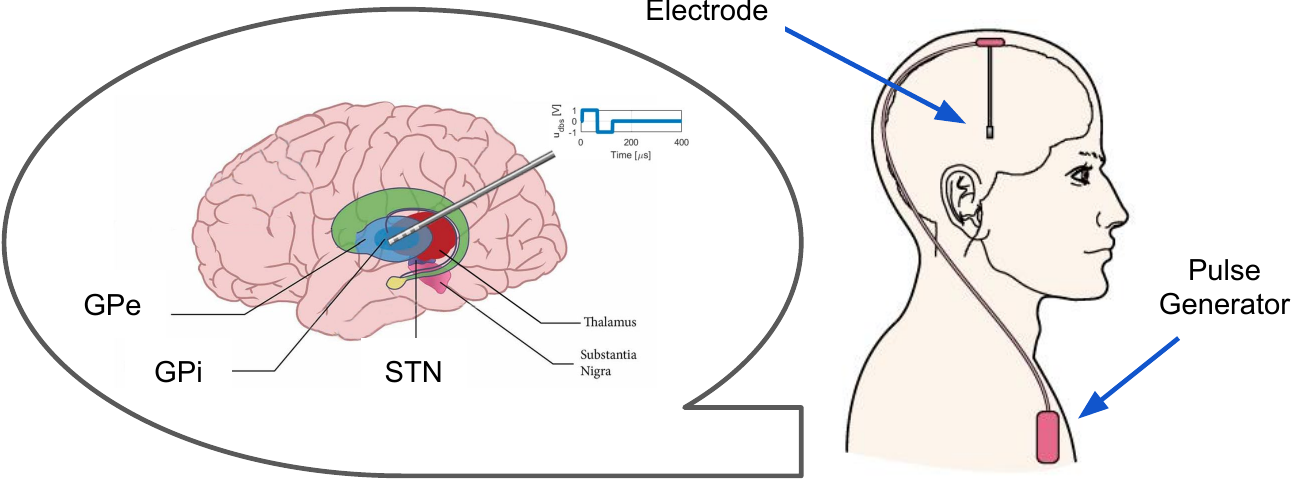}
    \caption{Deep brain stimulation: the implantable pulse generator is placed in the patient’s chest; electrodes that can record local field potentials (LFPs) and
deliver stimulation are positioned in the basal ganglia (BG) to stimulate the subthalamic
nucleus or the internal segment of the globus~pallidus (GPi).
    }  
    \label{fig:dbs}
\end{figure}

Millions of individuals in the U.S. suffer from Parkinson's disease (PD), a neurodegenerative disorder causing motor symptoms such as tremors, muscle stiffness, and bradykinesia~\cite{Marras2018}. Deep brain stimulation (DBS) has become widely used to treat motor symptoms by delivering electric pulses to the basal ganglia (BG) region of the brain through implantable devices \cite{DBSPD2003, randomDBS, pallidalDBS2010, DBSPD2012} illustrated in \Cref{fig:dbs}. DBS system consists of two main components: the electrode and the pulse generator. The electrode is a thin and insulated wire implanted in the brain with its tip positioned within the BG region. The pulse generator is usually placed under the skin near the collarbone or implanted closer to chest or abdomen. Two components are connected with an insulated wire passing under the skin of the head, neck, and shoulder so that
the electrical impulses can be sent from the pulse generator, up along the extension wire and the electrode, and into the brain for treatment.

DBS can significantly improve patients' daily life by alleviating PD symptoms; however, existing commercial DBS devices can only provide stimuli with pre-determined and fixed parameters (\textit{e.g.,} pulse frequency and amplitude)~--~\textit{i.e.}, continuous DBS (cDBS). To facilitate desirable therapeutic outcomes, the process of determining the parameters is often time-consuming because the parameters are usually determined by trial-and-error over multiple clinical visits \cite{Pineau2009}. In addition, stimulation with constant high frequency and amplitude significantly shortens the battery life of the implantable device and can result in serious side effects \cite{adbs2016}. Therefore, there has been a notable surge in research focused on advancing the automation of parameter selection for DBS, especially feedback-based stimulation controllers.

Existing research has primarily focused on developing adaptive DBS (aDBS) techniques, which automatically adjust stimulation parameters using various electrophysiological biomarkers as feedback signals \cite{adbs2016, adbsChallenge2016, adbsPD2016, little2013, little2016,gao_iccps22,gao_iccps23,schmidt_brain23}; specifically, local field potentials (LFPs) from the BG, internal electroencephalography (iEEG), and data from wearable devices such as electromyography and accelerometers with predefined thresholds established by physicians based on trial data. While the aforementioned aDBS approaches show promise in reducing energy consumption and mitigating stimulation-related side effects \cite{little2016, habets2018,gao_iccps22,gao_iccps23,schmidt_brain23}, the configuration of aDBS devices to optimize the balance between stimulation efficacy and battery efficiency remains a labor-intensive task. To reduce these substantial efforts, a distributed closed-loop neuromodulation architecture designed for the automated tuning of Proportional Integral (PI) controllers in DBS, leveraging Bayesian optimization is further introduced in~\cite{parisa2022}.

Recent research has explored the application of reinforcement learning (RL) to devise closed-loop controllers for aDBS in the context of PD. In particular, the studies conducted by \cite{seizure2017} introduce an approach where EEG and LFP signals are employed to define the state space within the RL framework. They employ fitted Q-iteration to synthesize RL control policies. These policies are geared toward selecting stimulation frequencies that aim to enhance energy efficiency. The work presented in \cite{qitong2020}, and extended in~\cite{gao_iccps22,gao_iccps23} utilizes deep actor-critic RL to craft personalized stimulation patterns tailored to each patient, which improves both energy efficiency and stimulation efficacy.

Compared with RL methods, contextual multi-armed bandit (CMAB) algorithms are more sample-efficient, which can better facilitate real-world DBS applications as data collection with human participants can be costly~\cite{analy_bandit, hedging_rl}. Moreover, lower computational resources are required for training and evaluating CMAB policies in general, facilitating better compatibility with the latest generation of embedded DBS systems, which do not provide the functionalities and bandwidth needed by executing full RL policies in real-time~\cite{commercial}. 

In this work, we introduce a CMAB approach to adapt the stimulation frequency of DBS, in response to the contexts defined as the beta-band (13-35 HZ) power spectral density ($P_\beta$) \cite{beta_power} of the LFP signals collected from the BG~\cite{bgm2012, jovanov_iccps18}. Moreover, the bandit arms represent the (discretized) stimulation frequency from the range of 0 Hz (\textit{i.e., turn off DBS}) to 180 Hz (\textit{i.e., cDBS}). Specifically, we propose an algorithm called $\epsilon$-Neural Thompson Sampling ($\epsilon$-NeuralTS), which blends deep neural networks with Thompson Sampling (TS) \cite{thompson}. It optimizes neural network approximators over a Bayesian objective, to estimate the posterior return distribution, normally parametrized as Gaussian, capturing the expected return of each arm with confidence, \textit{e.g.}, conditional (co-)variances to quantify the level of uncertainty. 
In addition, the arms are selected greedily by directly maximizing the expected return (\textit{i.e.}, the expectation of the posterior distribution) with probability $1-\epsilon$, or to sample from the posterior with probability $\epsilon$. Here, $\epsilon$ is hyper-parameter that helps balance exploration and exploitation. 

In what follows, a computational Basal Ganglia Model (BGM) \cite{qitong2020} is used as the testbed for training and evaluation of the CMAB policies, where $P_{\beta}$ and Error Index (EI) \cite{error_index} are considered as the Quality-of-Control
(QoC) metrics. We conduct comprehensive hyper-parameter tuning over the $\epsilon$ value and the reward function of CMAB method followed by the comparison with existing approaches. The results show that our method outperforms both existing cDBS methods and vanilla TS methods, in terms of the two metrics above as well as energy/time efficiency and robustness.


The main contributions of this work are:
\begin{enumerate}
    \item We re-formulate the aDBS problem into CMAB, where the interactions between the BG and the CMAB policy pertain to an environment with pre-defined feature contexts, action space, and reward functions.
    
    \item We propose a novel $\epsilon$-NeuralTS algorithm that is suitable for deployment over the latest generation of embedded DBS systems~\cite{commercial}. Moreover, it can trade off exploration and exploitation during training, leading to improved sample efficiency. As a result, it lays out the foundation of next-generation aDBS frameworks.

    \item We successfully demonstrate that our method outperforms several baselines, including both existing CMAB baselines and cDBS from the perspective of task performance and real-world scenarios.
\end{enumerate}
The paper is organized as follows.~\Cref{sec:preliminaries} introduces a Basal Ganglia Model (BGM) of the brain, the QoC metrics used to evaluate DBS control performance and the background of existing algorithms from MAB to CMAB. Our problem statement is formulated with the adaption of CMAB to DBS in~\Cref{sec:formulation}. We introduce our proposed $\epsilon$-NeuralTS in~\Cref{sec:method}. The experiments and analyses of the results are elaborated in~\Cref{sec:experiments}. This work is concluded with possible future extensions in~\Cref{sec:conclusion}.

\begin{figure}
    \centering
    \includegraphics[width = .4\textwidth]{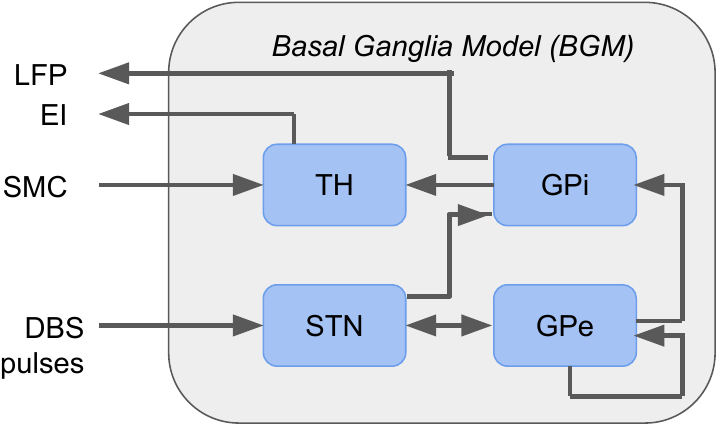}
    \caption{An illustration of the computational brain model. The DBS stimulation is deployed to the subthalamic nucleus (STN), propagating to the other sub-regions. Error index (EI) is computed with the activations passing from sensorimotor cortex (SMC) to thalamus (TH). 
    }
    
    \label{fig:bgm}
\end{figure}

\section{Preliminaries}\label{sec:preliminaries}
In this section, we start by describing the BGM with a formal definition of the QoC metrics the will be utilized for evaluating DBS control performance. We refer readers to~\cite{bgm2012,jovanov_iccps18} for an in-depth review of the model. We then introduce the background of Multi-armed Bandit (MAB) and extend it to contextual bandit settings. 

\subsection{Computational BGM}
The Basal Ganglia (BG) is a prominent cerebral region composed of three principal sub-regions, namely, the subthalamic nucleus (STN), globus pallidus pars externa (GPe), and globus pallidus pars interna (GPi). To comprehensively capture and quantify the manifestations of Parkinson's disease (PD), it is imperative to include not only these sub-regions but also the thalamic region (TH) and the sensory-motor cortex (SMC) inputs within the PD-specific brain model, as illustrated in~\Cref{fig:bgm}.

Supposing that there exists $n$ neurons in each sub-region, the state from the computational BGM at each time $t$ can be succinctly represented as a vector denoting electrical potential, as delineated below:
\begin{align}\label{eq:potential}
    \bm{v}^q(t) = [\nu^q_1, ...,  \nu^q_n ];
\end{align}
here, $\nu_j^q$ denotes the value of the $j^{th}$ neuron with the corresponding sub-region $q \in \{STN, GPe, GPi, TH\}$. The initial states of these neurons are considered model parameters that are stochastically determined in our experimental setup. Neurons are interconnected through chemical synapses, forming BGM structure illustrated in \Cref{fig:bgm}. The neural activation of each neuron at time $t$ is captured by binary events $a_j^q \in \{0,1\}$, which occur when the neuron's electrical potential $\nu_j^q$ exceeds a predefined threshold $h_j^q$, defined as
\begin{align}
    a_j^q(t) = \mathbb{I}\bigg([v_j^q(t) > h_j^q] \wedge [\exists\delta, \forall \epsilon \in (0,\delta), v_j^q(t-\epsilon) < h_j^q] \bigg).
\end{align}

We now formally define the two QoC metrics (\textit{i.e., }$P_{\beta}$ and EI) in order to evaluate the efficacy of DBS. 

\subsubsection{Error Index (EI)}
EI is defined as the portion of erroneous TH neuron activations in response to SMC inputs\footnote{Healthy brains could also respond to $SMC_\tau$ erroneously with a low probability
($< 0.1\%$).} $SMC_\tau$ at $t = \tau$. Specifically, $SMC_\tau$ can modulate TH neuron potentials and is expected to activate all TH neurons exactly once within a time window of $25 ms$ in healthy brains. In contrast, in the context of Parkinson's disease (PD), no such response or activation should be present within the $25 ms$ window immediately following the reception of an SMC input. Formally, EI is defined as 
\begin{align}
    EI(t) = \frac{\sum_{i=1}^n \sum_{t=t_0}^t a_i^{TH, err}(t)}{n \Bigl|SMC_{\tau}|_{t_0}^t\Bigr|},
\end{align}
where $a_i^{TH, err}(t) = 1 (\text{or } 0)$ indicates an erroneous (or correct) TH neuron activation at time $t$. Intuitively, every neuron in TH, and $\Bigl|SMC_{\tau}|_{t_0}^t\Bigr|$ is the cumulative number of SMC inputs received between the initial time $t_0$ and the current step $t$. Note that EI is bounded to the range $[0,1]$ because EI is defined as a ratio. The goal for the DBS controller is to maintain EI as low as possible.

\subsubsection{Beta-band Power Spectral Density ($P_\beta$)} In a PD brain, the GPi region exhibits pathological oscillations of neurons at frequencies within the $13 Hz - 35 Hz$ band (\textit{i.e., }beta band), which do not exist in a healthy brain. $P_{\beta}$ is defined as
\begin{align}
    P_{\beta j}^{GPi} =  \int_{\omega=2\pi\cdot13Hz}^{2\pi\cdot35Hz} P_j^{GPi}(\omega)d\omega,
\end{align}
where $P_j^{GPi}(\omega)$ is the single-sided power spectral density of the $j^{th}$ neuron's potential in the GPi region. Therefore, the beta band power for the entire region with $n$ neurons can be computed as
\begin{align}
    P_{\beta} = \frac{1}{n} \sum_{j=1}^n P_{\beta j}^{GPi}.
\end{align}
Note that EI can directly distinguish healthy brains from PD ones, but is intractable to be obtained in the real world~\cite{qitong2020}. On the other hand, $P_{\beta}$ can be noisy and sometimes may not distinguish healthy and PD brains at the same level of EI \cite{qitong2020}, $P_{\beta}$ can be easily obtained by typical DBS systems in clinical practice~\cite{commercial}; thus, it also serves as an imperative biomarker to quantify the severity of PD. The details of how we evaluate the feasibility of $P_{\beta}$ will be described in \Cref{sec:beta_power_eval}. 

\subsection{Multi-armed Bandit (MAB)}
We first introduce the basics of MAB, which is necessary for reviewing the preliminaries of CMAB. The MAB problem is a sequential game between a bandit learner and the environment. The game is played over $T$ rounds\footnote{In this work, we alternatively use \textbf{round}, \textbf{time}, and \textbf{time step} according to the corresponding context but with the same meaning.}, where $T$ is a positive integer called the horizon. At each time $t \in [0,T]$, the bandit learner first chooses an action $a_t$ from a given set $\mathcal{A}$, and then the environment receives the corresponding reward $R_t \in \mathbb{R}$. Actions are often called \textit{arms}, so $K$-armed bandits indicate that the cardinality of $\mathcal{A}$ is $K$. The bandit learner should choose $a_t$ depending on the history $D = (a_1, R_1, ..., a_{t-1}, R_{t-1})$. The common objective of the bandit learner is to learn a policy, which is a mapping from history to the next action, to maximize the cumulative reward over all $T$ rounds. The final performance is evaluated by \textit{regret}, which is defined as follows.

\begin{definition}[Regret]~\cite{bandit}
    The regret of the bandit learner with respect to a policy $\pi$  is the difference between the total expected reward obtained by using policy $\pi$ for $T$ rounds and the total expected reward collected by the bandit learner over $T$ rounds. The regret relative to a set of policies $\Pi$ is the maximum regret relative to any policy $\pi \in \Pi$ in the set.
\end{definition}

The main challenge in the bandit problem is addressing the exploitation-exploration trade-off, which targets reaching a subtle balance between following the myopically better arm and choosing an under-sampled worse arm. Existing algorithms for maximizing the cumulative
reward in bandits problems mainly follow either one of the following two algorithmic frameworks~--~upper confidence bound (UCB) and Thompson sampling (TS), as introduced in \Cref{sec:UCB} and \Cref{sec:TS}, respectively. 

\subsubsection{Upper Confidence Bound}\label{sec:UCB}
The UCB algorithm leverages the principle of optimism in the face of uncertainty. The optimal principle means using the data observed so far to assign to each arm a value (\textit{i.e.,} UCB which with high probability is an overestimate of the unknown mean). Assuming the upper confidence bound assigned to the optimal arm is indeed an overestimate, then another arm can only be played if its UCB is larger than that of the optimal arm, which in turn is larger than the mean of the optimal arm. Then the additional data provided by playing a suboptimal arm means that the UCB for this arm will eventually fall below that of the optimal arm. 

UCB is defined formally as follows. Let $(R_t)^T_{t = 1}$ be a sequence of independent 1-subgaussian random variable with mean $\mu$ and $\hat{\mu} = \frac{1}{n} \sum_{t=1}^T R_t$. Then
\begin{align}
    \mathbb{P}\bigg(\mu \geq \hat{\mu} + \sqrt{\frac{2 \log (1/\delta)}{n}}\bigg) \leq \delta, \forall \delta \in (0,1).
\end{align}
When considering its options in time $t$, the bandit learner has observed $T_k(t-1)$ samples from arm $k$ and received rewards from that arm with an empirical mean of $\hat{\mu}_k(t-1)$. Then a reasonable candidate for the unknown mean of the $k^{th}$ arm as large as plausibly possible is
\begin{align} 
    UCB_k(t-1, \delta)=
    \begin{cases}
        \infty & \text{if } T_k(t-1) = 0 \\
        \hat{\mu}_k(t-1) + G & \text{otherwise,} 
    \end{cases}			
\end{align}
where $G = \sqrt{\frac{2 \log (1/\delta)}{(T_k(t-1))}} $. The index of UCB is the sum of the empirical mean of rewards experienced so far and the exploration bonus (\textit{i.e.,} confidence width). Note that the exploration bonus has different versions according to different types of UCB algorithms, such as asymptotic optimality and minimax optimality. The high-level structure of the UCB-based algorithm is to start with the inputs of the number of arms $K$ and the error probability $\delta$. For each time $t \in [T]$, the bandits leaner choose arm $a_t = \argmax_i UCB_k(t-1, \delta)$ right before observing reward $R_t$ and updating UCB. Following this algorithm, the bandit learner explores arms more often if they are (a) promising because $\hat{\mu}_k(t-1)$ is large or (b) not well explored because $T_k(t-1)$ is small.

\subsubsection{Thompson sampling}\label{sec:TS}

TS, also called, posterior sampling, tackles MAB problems using a Bayesian approach. TS maintains a probability distribution for each arm's expected reward, representing their uncertainty about the true reward distribution. Given the set of history $D$, TS approach aims to learn the parameter $\bm{\theta}$ of the true reward distribution. TS starts with a prior distribution and a posterior distribution of $\theta_k$ for each arm $k \in [0, K-1]$, if we view $\bm{\theta}$ as the concatenation of $\theta_k$. This posterior distribution can be acquired by the Bayes rule, $P(\bm{\theta}|D) \propto  P(R_t|a_t, \bm{\theta})P(\bm{\theta})$, where $P(R_t|a_t, \bm{\theta})$ is a parametric likelihood function. 

In the vanilla TS, the algorithm samples from the corresponding posterior distributions $\theta_k(t)$ for all  $k \in [0, K-1]$, and selects the best arm $a_t = \argmax_{k \in [0, K-1]} \theta_k (t)$ right before observing reward $R_t$ and updating the corresponding posterior distribution. We refer to the details of TS in~\cite{bandit}.

\subsection{Contextual Bandits}\label{sec:CB}
Contextual bandits are a wide class of sequential decision problems, where the bandit learner makes the decision based on an observation of an action set consisting of feature vectors as contexts for different actions. In particular, at time $t \in [0,T]$, the bandit learner observes the context $\bm{x}$ consisting of $K$ context vectors $\{\bm{x}_{t,k} \in \mathbb{R}^d | k \in [0,K-1]\}$. The bandit learner then selects an action $a_t \in \mathcal{A}$ and receives the corresponding reward $R_{t,a_t} = h(\bm{x}_{t,a_t}, \bm{\theta}) + \xi_t$, where $h: \mathbb{R}^d \rightarrow \mathbb{R}$ and $\bm{\theta} \in \mathbb{R}^d$ is an unknown weight parameter for bandit learner; $\xi_t \in \mathbb{R}$ is a random noise incurred in the observation, which is standard in the stochastic bandit literature \cite{LinTS, foundationML}. To simplify our notation, we can assume that the reward is independent of the time $t$ and the noise $\xi_t$ can be ignored. We can further formulate our reward function with the whole context $\bm{x}$ expressed as $R = h(\bm{x}, \bm{\theta})$ by dropping the subscripts. For instance, we have $h(\bm{x}, \bm{\theta}) = \bm{x}^\top \bm{\theta}$ in linear contextual bandits \cite{NIPS2011_e1d5be1c, LinTS, LinUCB}, and $h(\bm{x}, \bm{\theta}) = \mu(\bm{x}^\top \bm{\theta})$ for generalized linear bandits \cite{NIPS2010_c2626d85, random2020bandit, Li2017ProvablyOA, Ding2020AnEA}, where $\mu(\cdot)$ is a link function. Our work aligns with the neural contextual bandits \cite{Riquelme2018, NeuralTS, NeuralUCB, Xu2020NeuralCB} so that $h(\bm{x}, \bm{\theta})$ is a neural network, where $\bm{\theta}$ is the concatenation of all weight parameters and $\bm{x}$ is the input.

Within the realm of contextual bandit problems, algorithms grounded in Optimism in the Face of Uncertainty (OFU) are often required to solve a bi-linear optimization problem, which makes them computationally expensive to implement outside of simple problems despite their stronger theoretical guarantees. In contrast, Thompson Sampling (TS) algorithms offer a more computationally efficient alternative. These methods only require solving a linear optimization problem on the set of available arms. This efficiency stems from the fact that the inherent uncertainty encapsulated within the posterior distribution naturally accommodates exploration in the parameter space. Moreover, it is noteworthy that TS has been observed to be empirically competitive with or even superior to OFU-based algorithms in practical scenarios~\cite{empirical_ts}.

 
\section{Problem Formulation}\label{sec:formulation}
We formulate a $K$-armed CMAB problem for selecting the frequency of the stimulus for PD in DBS with $P_\beta$ as context inputs.  Specifically,  the context features $s_t$ at a discrete round, which can be defined as a sequence of $P_\beta$ at a fixed rate, $m \in \mathbb{Z}^{+}$, over a window of size $T_w$. \textit{i.e., }
\begin{align}\label{eq:state}
    s_t = [\beta_{(t)}, \beta_{(t+m)},\beta_{(t+2m)},..., \beta_{(t+T_w -m)}],
\end{align}
where $\beta_{(\cdot)}$ represents $P_\beta$ evaluated at $l = T_w / m$ number of equally-spaced intervals within the window. 
The bandit learner can select its action in time $t$ as $a_t$ from $K$ arms, where $K = 13$ in our problem setting. We limit the maximum stimulus frequency to $180 Hz$ in the computational BGM. 

To have a better action mapping strategy, according to each arm $k \in [0, K-1]$, we can have  $F = 15k$ ($\textit{e.g.,}$ when $k=12$, the stimulus frequency achieves $180 Hz$). Then the selected arm $a_t$ can be mapped back to the action space for the BGM. We change the stimulation frequency every $T_w$ steps, so the mapped action $u_t$ that the bandit learner can take at time $t$ is 
\begin{align}
    u_t &=  [u_{(t)}, u_{(t+m)},u_{(t+2m)},..., u_{(t+T_w -m)} ],\\
    u_{(t+j)} &= 
    \begin{cases}
        1 & \text{if a pulse is triggered at time $t+j$} \\
        0 & \text{otherwise,} 
    \end{cases}
\end{align}
where $j \in [0, T_w -m)$. Finally, we define our reward function as $R_{t,k} = - \Bar{s}_{t+1}-C \cdot a_t$, where $\Bar{s}_{t+1}$ is the mean of the whole vector of $s_{t+1}$ in \eqref{eq:state} and $C \cdot a_t$ is the selected action from the bandit learner multiplied by a constant coefficient $C \in \mathbb{R}$. Again, here the value of $a_t$ is $k \in [0, K-1]$. 

Therefore,  $C \cdot a_t$ can be viewed as a penalty on the frequency value, which can encourage the bandit learner to reduce the $P_\beta$ defined in $s_{t+1}$ as well as to consume less energy, resulting in energy efficiency and relatively mild side effects, which is highly relevant to safety issues and therapy effectiveness~\cite{SaferlHsu2022, neuroweaver}. Typically the goal in CMAB is to choose actions that maximize the cumulative reward over $T$ steps, which is equivalent to minimizing the cumulative regret $r(T)$, defined as the difference between the maximum possible context-dependent reward and the actually received~reward
\begin{align}
    r(T) = \mathbb{E}\bigg[\sum_{t=1}^T (R_{t, k^*}^* -  R_{t, k})\bigg],
\end{align}
where $R_{t, k^*}^*$ is the reward with optimal action $a_t = k^*$ and $k^* \in \argmax_{k} \mathbb{E}[R_{t, k}]$. However, note that EI is the oracle (ground truth) to evaluate the severity of PD symptoms and the optimal value of EI can be minimized to be closer to $0$. With the property of EI, we can introduce EI as our regret for the final evaluation\footnote{Note that the EI is not involved in the reward function and the context feature during~learning.} and we quantify the task performance of different algorithms by comparing each cumulative regret.

Finally, besides evaluating the task performance, our goal is to also extract the energy consumption component from the reward function as the evaluation of energy efficiency.

 \begin{algorithm}[!t]
	\caption{$\epsilon$-Neural Thompson Sampling ($\epsilon$-NeuralTS))}
	\begin{algorithmic}[1]
		\STATE \textbf{Input}: number of rounds $T$, exploration variance $\nu$, initialized weight of neural network $\bm{\theta}_0$ with network width $m$, regularization parameter $\lambda$, exploration probability $\epsilon$
            \STATE $\bm{U}_0 = \lambda \bm{I}$
		
		\FOR{$t =1,...,T$}
			\FOR{$k =1,...,K$}
                    \STATE  {\small\begin{align*}
					R_{t,k} 
					\begin{cases}
						\sim \mathcal{N}(f(\bm{x}_{t,k}),\nu^2\sigma^2_{t,k}) & \text{w.p. } \epsilon \\
						=f(\bm{x}_{t,k}) & \text{w.p. } 1-\epsilon 
					\end{cases}
				\end{align*}}  \label{line:sample} 
			\ENDFOR
		
			\STATE Pull arm $a_t$ and receive reward $R_{t,a_t}$, where $a_t = \argmax_{k \in [0, K-1]} R_{t,k} $ \label{line:pull}
   \STATE Set $\bm{\theta}_t$ as the output of gradient descent for solving \eqref{eq:l2_loss} \label{line:theta}
   \STATE $\bm{U}_{t} =\bm{U}_{t-1} + g(\bm{x}_{t, a_t};\bm{\theta}_t) g(\bm{x}_{t, a_t};\bm{\theta}_t)^\top /m$\label{line:cov}
		\ENDFOR
	\end{algorithmic}
	\label{algo:eps_neuralTS}
\end{algorithm}

\section{$\epsilon$-NeuralTS}\label{sec:method}
NeuralTS is a CMAB method designed to harness the potential of deep neural networks for both exploration and exploitation \cite{NeuralTS}. Central to this algorithm is an innovative approach to modeling the posterior distribution of rewards. Specifically,  compared to the typical ways of implementing TS with neural network sampling the weight parameters, NeuralTS samples from the posterior distribution of the scalar reward with the mean determined by the neural network approximator and the variance constructed based on the neural tangent features associated with the corresponding neural network. Therefore, NeuralTS is simpler and more efficient because the number of parameters can be large in practice.

During learning, the reward function is unknown to the bandit learner. To estimate the unknown reward given a contextual vector $\bm{x}$, we build a fully connected neural network $f(\bm{x}, \bm{\theta})$ for approximation~\cite{NeuralTS}, defined recursively by
\begin{align*}
    f_1 = \bm{W}_1 \bm{x},
\end{align*}
\begin{align*}
    f_l = \bm{W}_l \text{ReLU}(f_{l-1}), 2 \leq l \leq L,
\end{align*}
\begin{align}
    f(\bm{x}, \bm{\theta}) = \sqrt{m}f_L,
\end{align}
where $\text{ReLU}(x)\coloneqq  \max \{x,0\}$, $m$ is the width of the neural network, and $\bm{W}_i$ denotes as the weight parameters of $i^{th}$ layer in the full neural network. Therefore, $\bm{\theta} = \big(\text{vec}(\bm{W}_1); ...;\text{vec}(\bm{W}_L)\big)$ is the collection of parameters of the whole neural network. Finally, $g(\bm{x};\bm{\theta}) = \nabla_{\bm{\theta}} f(\bm{x}, \bm{\theta})$ is the gradient of $f(\bm{x}, \bm{\theta})$ w.r.t $\bm{\theta}$.

We summarize our $\epsilon$-NeuralTS in \Cref{algo:eps_neuralTS}. We firstly input the number of rounds $T$, exploration variance $\nu >0$, initialized neural network, regularization parameter $\lambda$, and exploration probability $\epsilon$. Then we initialize a covariance matrix $\bm{U}_0 = \lambda \bm{I}$, where $\bm{I}$ is an identity matrix.

Inspired by the recent work $\epsilon$-TS \cite{eps_TS, Jin2024MATS} for non-contextual and weight parameter sampling, our novel  $\epsilon$-NeuralTS builds upon NeuralTS with $\epsilon$ exploring. Specifically, for each time $t \in [0, T]$, we estimate the reward for each arm $k \in [0, K-1]$. When selecting an arm, it only explores with sampling the reward from its posterior distribution with probability $\epsilon$ while the arm is played based on empirical mean rewards with probability $1-\epsilon$, where $\epsilon \in (0,1)$ is a user-defined parameter in Line \ref{line:sample} in \Cref{algo:eps_neuralTS}. Note that the $\sigma_{t,k}$ in $\mathcal{N}(f(\bm{x}_{t,k}),\nu^2\sigma^2_{t,k})$ is calculated by
\begin{align}\label{eq:sigma}
    \sigma^2_{t,k} = \lambda g^\top(\bm{x}_{t,k}; \bm{\theta}_{t-1})\bm{U}_{t-1}^{-1} g(\bm{x}_{t,k}; \bm{\theta}_{t-1})/m.
\end{align}

Therefore, $\epsilon$-NeuralTS can improve both sample and computational efficiency by reducing the number of calculations. Then the bandit learner pulls the arm with the maximum estimated reward in Line \ref{line:pull}.
Once the reward is observed, it
updates the posterior (Lines \ref{line:theta} \& \ref{line:cov}). The mean of the posterior distribution is set to the output of the
neural network, whose parameter is the solution to the following $l_2$-regularized square loss minimization problem:
\begin{align}\label{eq:l2_loss}
    \min_{\bm{\theta}}L(\bm{\theta}) = \sum_{i=1}^t [f(\bm{x}_{i,a_i}, \bm{\theta}) - R_{i, a_i}]^2/2 + m \lambda \|\bm{\theta} - \bm{\theta}_0\|_2^2/2,
\end{align}
where the regularization term centers at the randomly initialized network parameter $\bm{\theta}_0$.

\section{Experiments}\label{sec:experiments}
In this section, we evaluate our proposed $\epsilon$-NeuralTS against other contextual bandits algorithms and the controller with periodic stimulation patterns that are employed in \cite{Pineau2009}, \cite{adaptiveRL2008},\cite{seizure2017} over computational BGM. For a fair comparison, we set up the sampling duration $l = T_w = 2$ seconds for all contextual bandits algorithms, indicating that the effect of every arm will last within this duration. 

Recall that our context feature $s_t$ is defined as a sequence of $P_\beta$ sampled at a fixed rate $m$ over a window size $T_w$. Since the context feature is shared among all arms, we follow~\cite{random2020bandit, Riquelme2018} to construct context vectors $\bm{x}$ for different arms in the following way: given a context feature $s \in \mathbb{R}^d$ (\textit{i.e., } $d=T_w$), we transform it into $K$ contextual vectors $\bm{x} = [\bm{x}^{(1)}; ...; \bm{x}^{(K)}] \in \mathbb{R}^{Kd}$ (\textit{e.g., }$\bm{x}^{(1)} =(s, 0, ..., 0)$ and $\bm{x}^{(K)} = (0, ..., 0, s) $). We learn the parameters $\bm{\theta}$ of the neural network, discussed in \Cref{sec:method}, with \textbf{x} as inputs for our $\epsilon$-NeuralTS. Specifically, we build a fully connected neural network with a sequence of $3$ layers ($32$ neurons per layer) followed by the ReLU activation function.

All our experiments are run on Nvidia RTX A5000 with 24GB RAM. In particular, 
our experiments focused on the following tasks:
\begin{enumerate}
    \item To evaluate the feasibility of using $P_\beta$ as a PD biomarker during learning for context feature and reward function, we conduct an experiment to find the correlation between $P_\beta$ and EI.
    \item We tune the coefficient of the penalty term in the reward function and the $\epsilon$ value of $\epsilon$-NeuralTS.
    \item We compare our proposed $\epsilon$-NeuralTS against existing CMAB methods and classical periodic controllers. Note that we do not explicitly compare our methods with other aDBS approaches because our environment setups are mostly not the same. In addition, although our experiment shares similar computational BGM with \cite{qitong2020}, we only consider $P_\beta$ as the input state, which contains less information but be more realistic.
    \item We evaluate the impact of different $\epsilon$ on the running~time $\epsilon$-NeuralTS algorithm.
    \item Finally, we evaluate $\epsilon$-NeuralTS on the robustness to delayed rewards.
\end{enumerate}

\begin{figure}
    \centering
    \includegraphics[width = .4\textwidth]{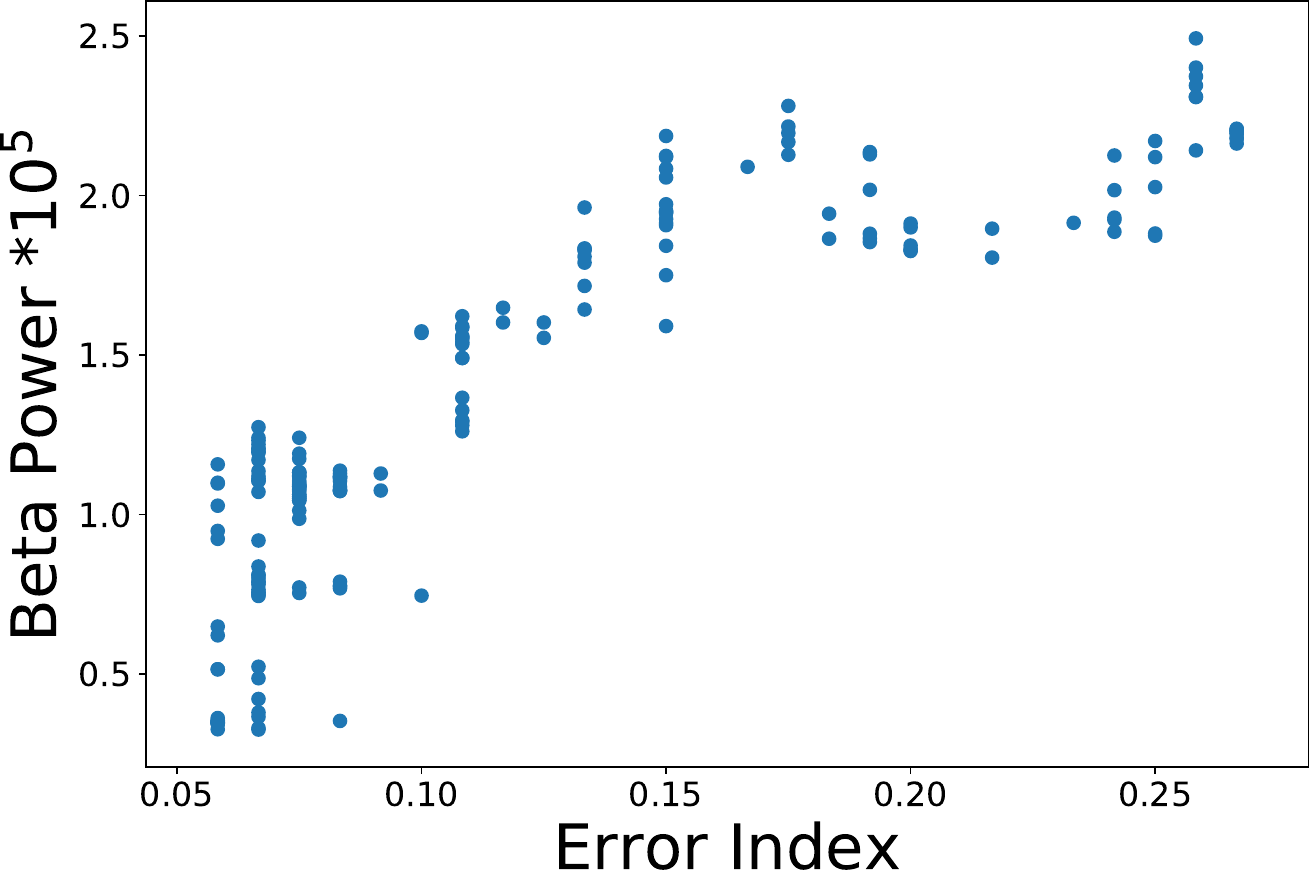}
    \caption{Correlation between two QoC (\textit{i.e., $P_\beta$ and EI}) with Pearson’s Correlation Coefficient: $0.866$. 
    }
    \label{fig:qoc}
\end{figure}
\subsection{Feasibility of $P_\beta$ as a PD Biomarker}\label{sec:beta_power_eval} To evaluate the feasibility of using $P_\beta$ as the PD biomarker during learning for context feature and reward function, we firstly experiment to find the correlation between $P_\beta$ and EI. In particular, we randomly deploy $9$ pulses within $200 ms$ on the computational BGM and collect the corresponding $P_\beta$ and EI. The distribution is shown in \Cref{fig:qoc} with Pearson’s Correlation Coefficient = $0.866$. With the high correlation between $P_\beta$ and EI (closer to $1.0$), $P_\beta$ can be seen as an indicator of PD symptoms with noises. Therefore, we adopt $P_\beta$ for the feature context and reward function as described in \Cref{sec:formulation}.

\begin{figure}[!t]
    \centering
    \includegraphics[width = .4\textwidth]{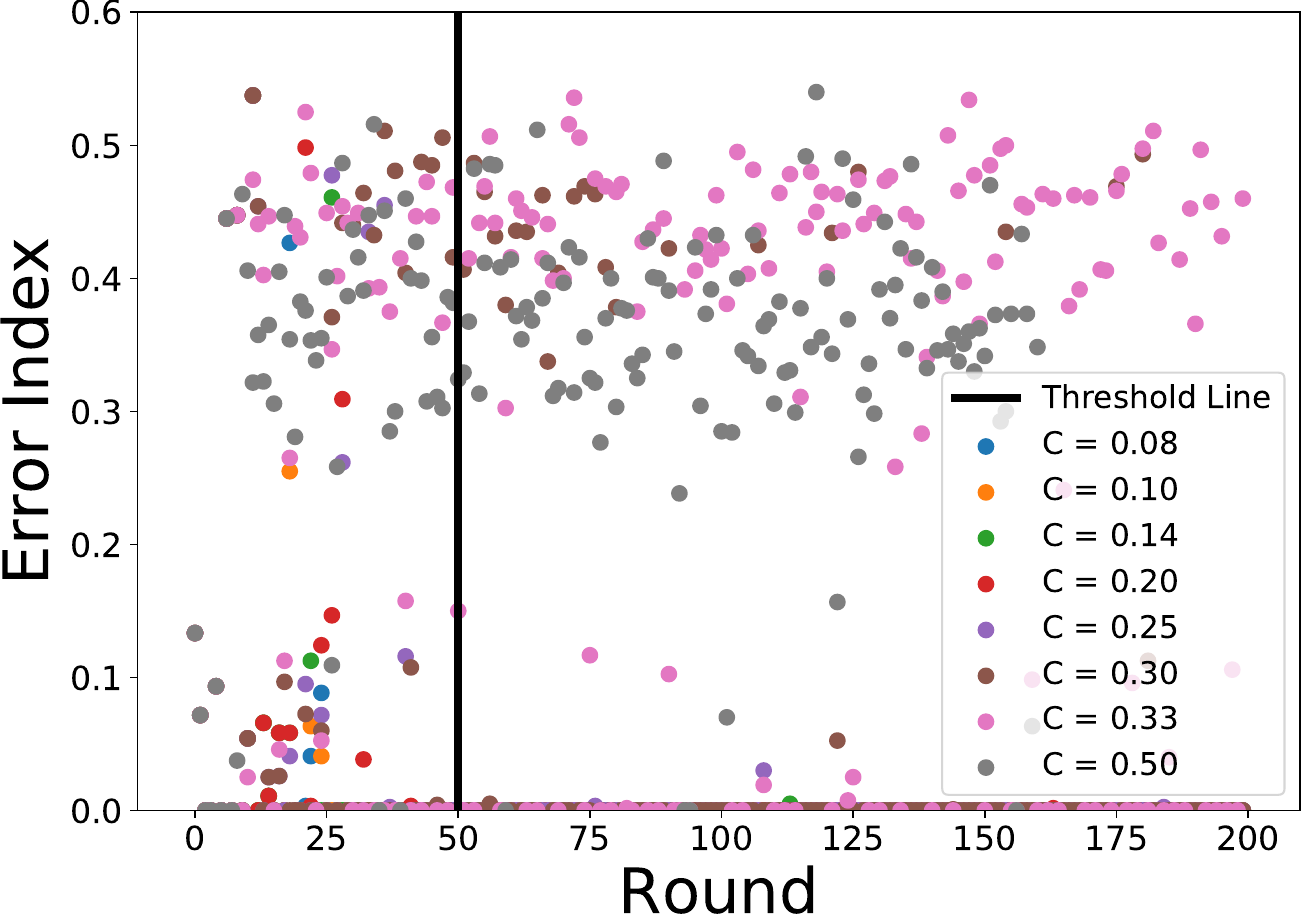}
    \caption{Learning curve with different penalty coefficients using NeuralTS (lower EI is better). 
    }
    \label{fig:reward_tune}
\end{figure}

\begin{table}[th!]
\caption{Average Frequency with different penalty coefficient after Threshold Line for NeuralTS.}
\begin{center}
        \label{table:coeff_freq}
    
        \resizebox{\columnwidth}{!}{
        \begin{tabular}{l | l l l l l l l l}
            \toprule
            \textbf{Penalty Coefficient $C$} & \textbf{0.08} & \textbf{0.10} & \textbf{0.14} & \textbf{0.20} & \textbf{0.25} & \textbf{0.30} & \textbf{0.33} & \textbf{0.50}\\
            \midrule
            Average Arm $k$       & 8.6         & 8.1               & 8.0 &9.0 & 8.9         & 7.9               & 4.1 &0.6 \\
            \bottomrule
        \end{tabular}}
        
   \end{center}
   
    \end{table}
\subsection{Hyper-parameter Tuning of Penalty Coefficient}
Recall that our reward function is defined as $R_{t,k} = - \Bar{s}_{t+1}-C \cdot a_t$, where $\Bar{s}_{t+1}$ is the mean of the whole vector of $s_{t+1}$ in \eqref{eq:state} and $C \cdot a_t$ is the penalty of the stimulus with higher frequency accompanied by a constant coefficient $C \in \mathbb{R}$. Note that in \cite{qitong2020}, the reward function is designed with $4$ discrete categories according to the values of $P_{\beta}$ and EI, which requires access to EI and more engineering work on deciding the reward values for $4$ different categories. Our penalty coefficient $C$ can be tuned within a smaller search space. Since the value of penalty coefficient $C$ will influence the reward function and EI can serve as the final evaluation, we aim to find a suitable $C$ so that the learned policy can maintain a low EI ($<0.1$) and lower stimulation frequency with less energy consumption and side effects. 

Intuitively, higher stimulation frequency can be more effective in suppressing PD symptoms and larger $C$ will discourage the policy from selecting a higher simulation frequency. Thus, a trade-off exists between the task and safety (\textit{e.g.,} side effects) performance. Task performance is our priority condition before we select the lowest average stimulation frequency.

To have a fair comparison, we consider our strong baseline NeuralTS~\cite{NeuralTS} for hyper-parameter tuning. \Cref{fig:reward_tune} shows the EI values with different coefficients $C = [0.08, 0.10, 0.14, 0.20, 0.25, 0.30, 0.33, 0.50]$. We observe that most of the settings have EI values smaller than $0.1$ after $t = 50$ rounds (\textit{i.e., } threshold line) while $C= [0.30, 0.33, 0.50]$ cannot converge even with longer rounds. We average the frequency after the threshold line for all the settings, resulting in $C = 0.14$ the lowest average frequency (\textit{i.e.,} stimulation frequency $F = 15 i$, where $k \in [0, K-1]$) with low EI ($< 0.1$) in \Cref{table:coeff_freq}; here, $K = 13$. Hence, we adopt $C = 0.14$ in our reward function for the remaining experiments. 

The results for the average frequency are reported~in~\Cref{table:coeff_freq}.

\begin{figure}
    \centering
    \subfigure[Reward]{
        \includegraphics[width =.22\textwidth]{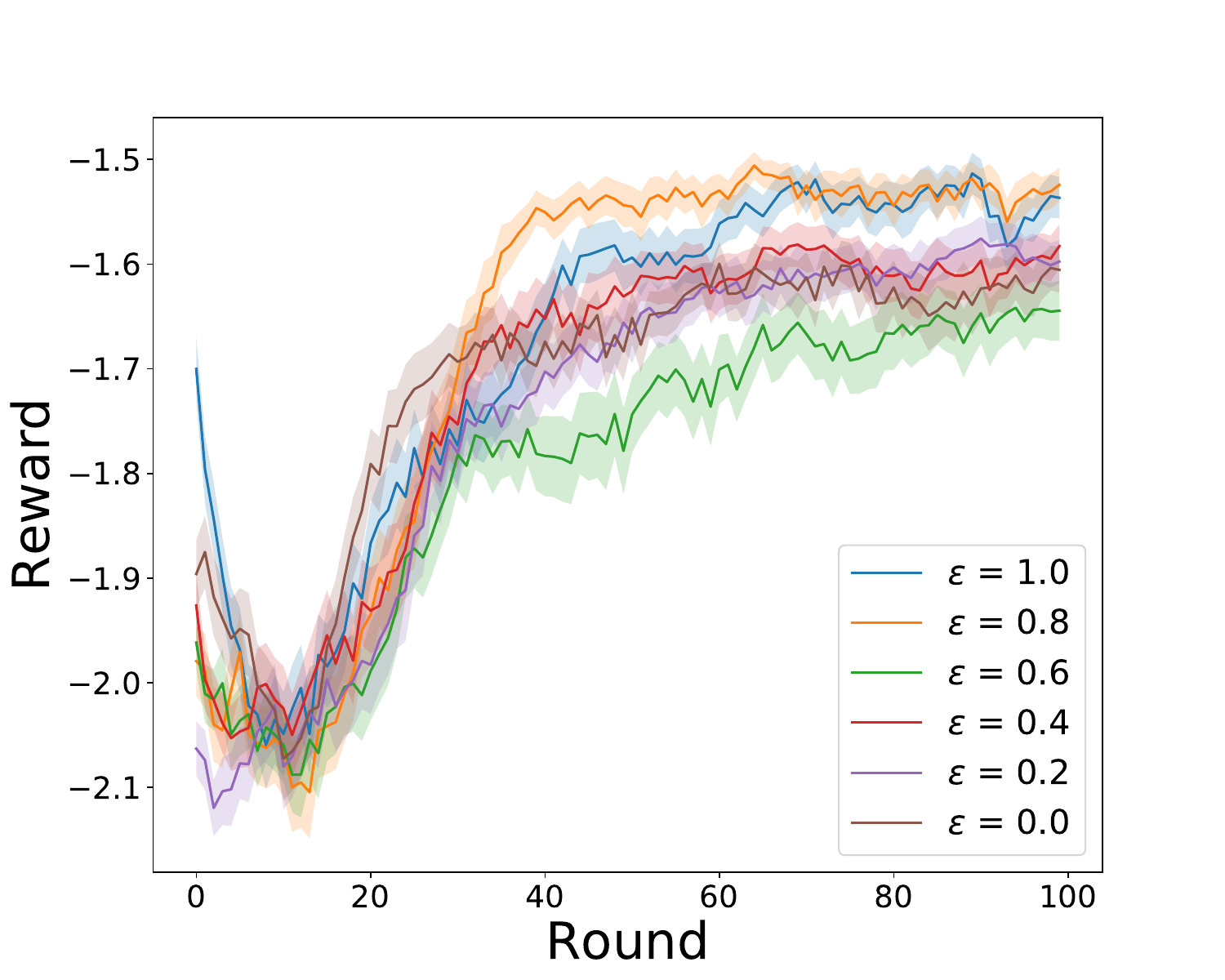}
      
    }
    \subfigure[Cumulative Regret]{
        \includegraphics[width =.22\textwidth]{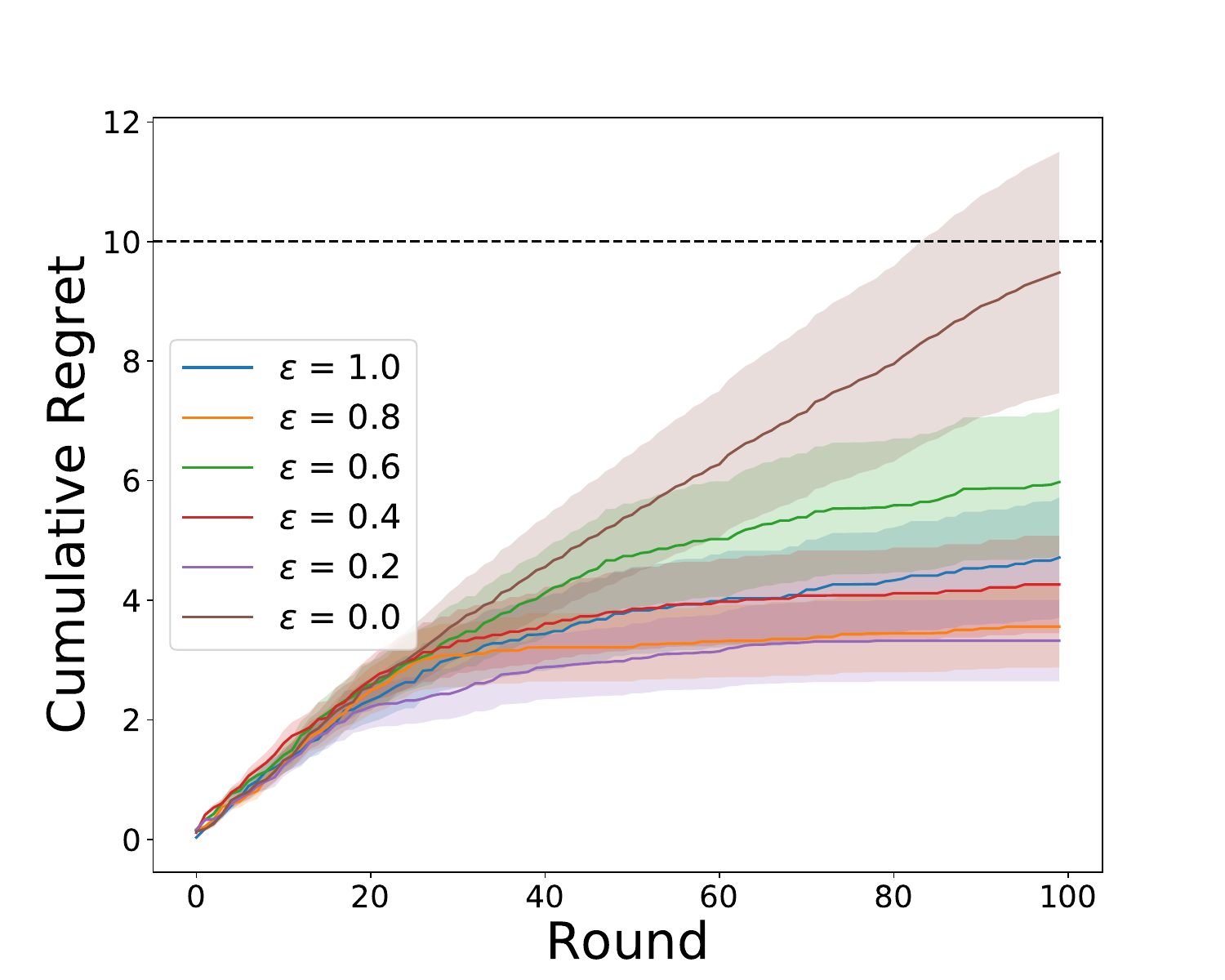}
        
     }
    \caption{Task Performance for $\epsilon$-NeuralTS with different $\epsilon$ averaged over $10$ seeds. Shaded areas denote the standard error: (a) task reward (higher is better), (b) cumulative regret (lower is better).}\label{fig:eps_neuralTS}
\end{figure}
\subsection{Hyper-parameter Tuning of $\epsilon$ for $\epsilon$-NeuralTS}
Before comparing with other CMAB methods, we investigate the optimal $\epsilon$ via our reward function, which has already been decided as a better trade-off between task performance (\textit{i.e., } $P_\beta$) and penalty on frequency value with coefficient. Note that NeuralTS can be viewed as the special case of $\epsilon$-NeuralTS with $\epsilon = 1.0$, so we also include it in comparison.

We conduct all the setups with different $\epsilon$ for $10$ random trials to represent $10$ different patients. We record the task reward and cumulative regret (\textit{i.e., }cumulative EI) with different $\epsilon$ in \Cref{fig:eps_neuralTS}. We observe that 
worse performance does happen with insufficient exploration when $\epsilon < 0.8$ in our task. However, when $\epsilon = 0.8$, we reduce by $20 \%$ the number of sampling and calculations with gradient descent as well as speed up the convergence with competitive performance compared with vanilla NeuralTS (\textit{i.e., }$\epsilon = 1.0$). On the other hand, we realize that the cumulative regret diverges when $\epsilon = 0.0$, which is reasonable due to the extreme imbalance between exploration and exploitation. However, $\epsilon = 0.2$ results in the minimum cumulative regret, showing that the suppression of PD with only $20 \%$ computational resource has minimum EI. 

Since we do not utilize any information from EI during learning, we believe that the mismatching between the reward and cumulative regret comes from two reasons: (\textit{i}) the correlation between $P_\beta$ and EI is not completely perfect, and (\textit{ii}) the designed reward function considers the regularization of the frequency values. This mismatching appears mainly with the extremely lower $\epsilon$, which is acceptable. In addition, $\epsilon$-NeuralTS with $\epsilon = 0.8$ still receives a lower cumulative regret versus NeuralTS, demonstrating the consistency of performance improvement on $\epsilon$-NeuralTS via less exploration. 

Overall, most of the settings except for $\epsilon = 0.0$ can converge below the dashed line, indicating that they can alleviate PD symptoms successfully because the dashed line is with cumulative regret $r_{(100)} = 0.1 (\text{EI}) \times 100 (\text{rounds})$; note that the healthy brains are with EI $< 0.1$. This observation 
effectively shows that we can consider less exploration carefully to achieve an improvement of NeuralTS. Therefore, we will further compare $\epsilon$-NeuralTS with $\epsilon = 0.8$ with other methods.

\begin{figure}[!t]
    \centering
    \subfigure[Reward]{
        \includegraphics[width =.22\textwidth]{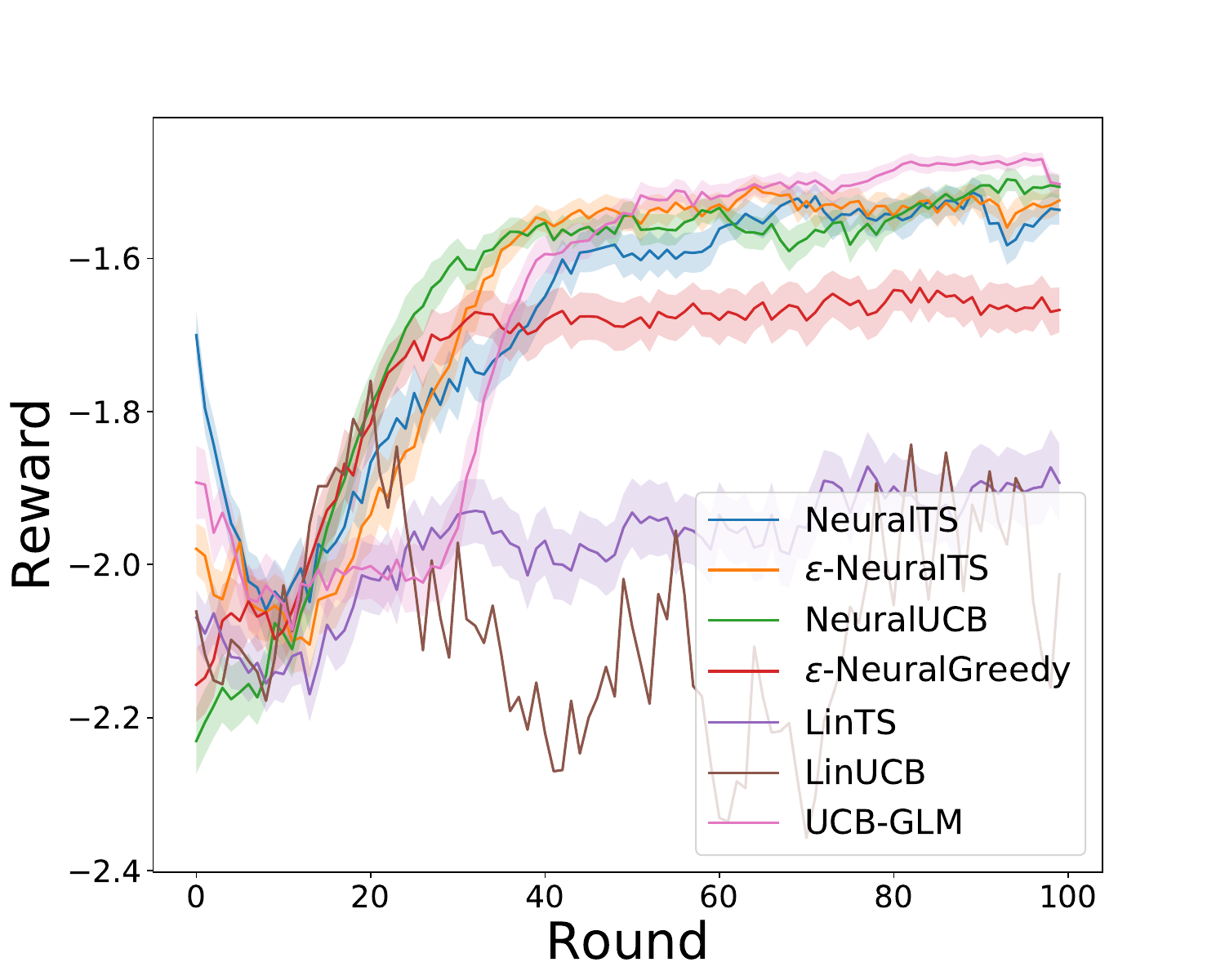}

    }
    \subfigure[Cumulative Regret]{
        \includegraphics[width =.22\textwidth]{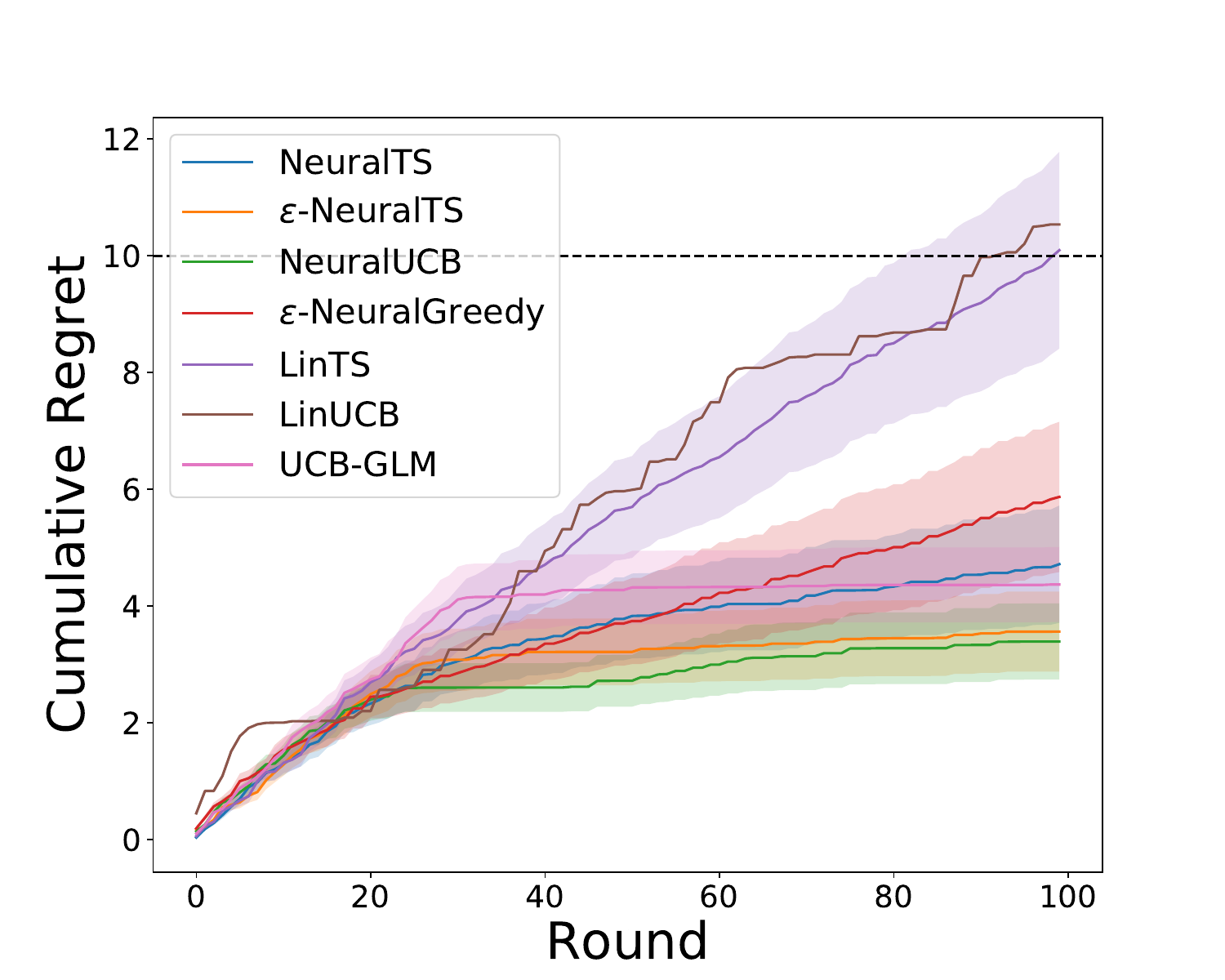}
        
     }
    \caption{Task Performance for $\epsilon$-NeuralTS against other methods averaged over $10$ seeds. Shaded areas denote the standard error: (a) task reward (higher is better), (b) cumulative regret (lower is better).}\label{fig:algo}
\end{figure}

\subsection{$\epsilon$-NeuralTS against CMAB Algorithms}\label{sec:exp_CMAB} In addition to considering the vanilla NeuralTS as our baseline, we also perform comparison to other existing CMAB algorithms, including linear bandit (\textit{e.g., } LinUCB \cite{LinUCB} and LinTS \cite{LinTS}), generalized linear bandits (\textit{e.g., }UCB-GLM \cite{UCBGLM}), and neural bandits (\textit{e.g., }NeuralUCB \cite{NeuralUCB} and Neural $\epsilon$-greedy \cite{NeuralGreedy}). Note that the $\epsilon$ in Neural $\epsilon$-greedy is not the same as the role of our $\epsilon$ in $\epsilon$-NeuralTS. Instead, $\epsilon$ in Neural $\epsilon$-greedy is for deriving a probability for randomly selecting action as exploration. Also, this probability for exploration will keep decreasing with increasing rounds so that the learning~converges. 

In \Cref{fig:algo}, we report the mean and the standard error of the cumulative regret of different algorithms over $10$ runs. We demonstrate that our $\epsilon$-NeuralTS with $\epsilon = 0.8$ is still competitive compared to the other algorithms. The results for linear bandit-based approaches (\textit{i.e., }LinUCB and LinTS) are the worst in both the reward and cumulative regret. UCB-GLM and vanilla NeuralTS perform well with high rewards while they receive relatively worse cumulative regrets, reflecting higher EI. We notice that NeuralUCB is the strongest baseline for our task, demonstrating similar reward and cumulative regret as $\epsilon$-NeuralTS with $\epsilon = 0.8$ does. 

We emphasize our contribution to improving the task performance of the vanilla NeuralTS using $\epsilon$-exploring strategy so that the branch of NeuralTS can be competitive with NeuralUCB in this task. Also, less exploration means that we reduce the risk of searching for unknown scenarios, which is critical for medical devices and procedures in the real~world.


\begin{figure}[!t]
    \centering
    \includegraphics[width = .4\textwidth]{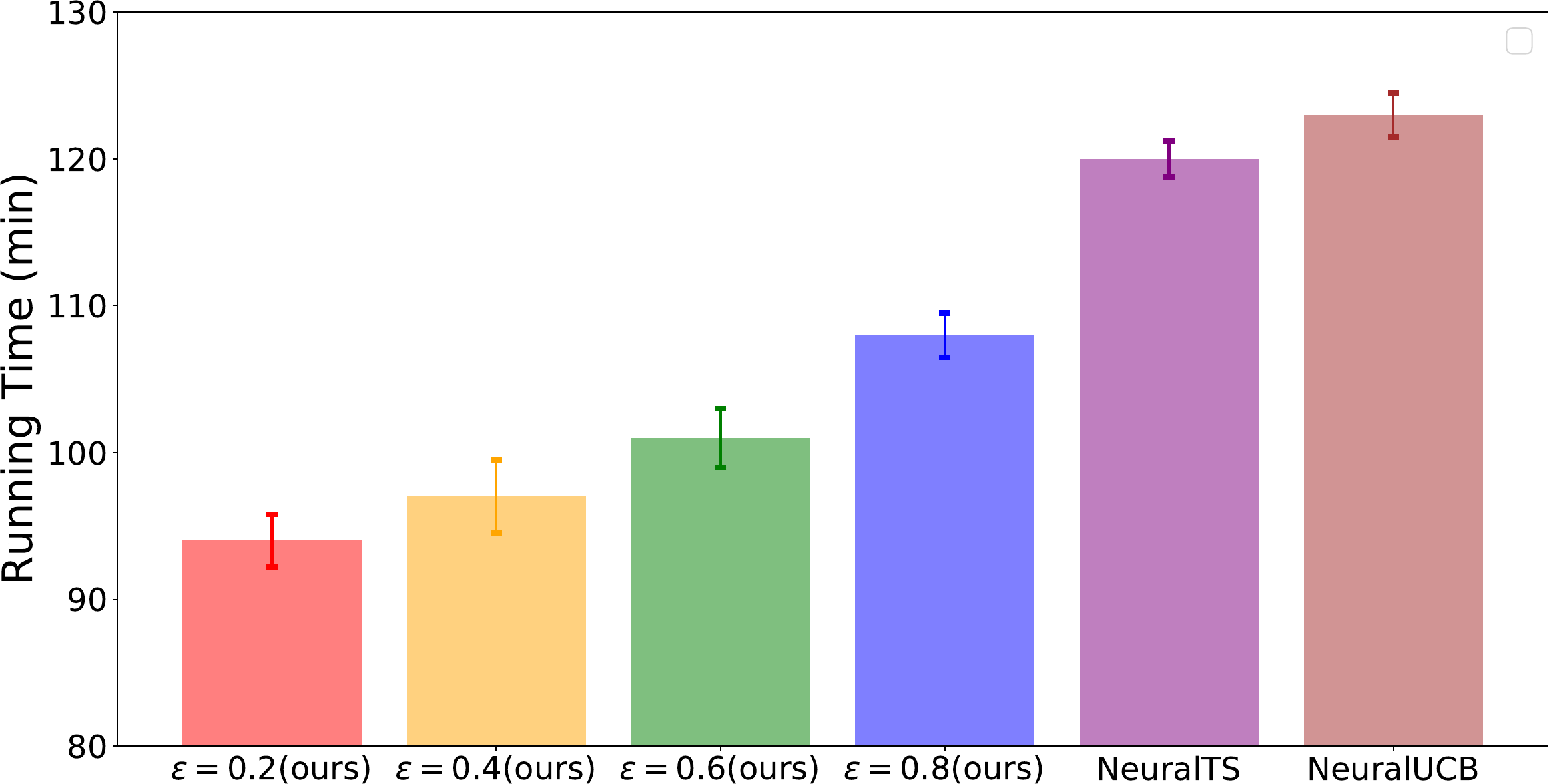}
    \caption{Running time comparison of different $\epsilon$ for our $\epsilon$-NeuralTS versus NeuralTS and NeuralUCB. Each trial takes $100$ rounds of interaction with the computational BGM.
    }
    \label{fig:running_time}
\end{figure}

\subsection{Relative Computation Time of $\epsilon$-NeuralTS}
Since the sampling with exploration in the vanilla NeuralTS requires computation including additional taking gradient descent w.r.t $\bm{\theta}$: $g(\bm{x};\bm{\theta}) = \nabla_{\bm{\theta}} f(\bm{x}, \bm{\theta})$, which is described in \eqref{eq:sigma}, exploration reduction can also reduce the computation time during learning. We compare the running time of each trial under different $\epsilon$ against the vanilla NeuralTS and NeuralUCB; the results are summarized in \Cref{fig:running_time}. Specifically, each trial takes $100$ rounds of interaction with the computational BGM. We notice that NeuralTS runtime is smaller than NeuralUCB with the same number of rounds, which is consistent with the mathematical perspectives mentioned in \Cref{sec:CB}. 

Also, reducing $\epsilon$, the $\epsilon$-NeuralTS run time will decrease. Since our best setting in the task of suppressing PD symptoms is with $\epsilon = 0.8$. We report that the running time is about $10 \%$ less than the standard NeuralTS with $\epsilon = 1.0$. Note that the interaction with computational BGM also occupies a huge portion of running time, which is the same for all algorithms.  

\begin{figure}[!t]
    \centering
    \includegraphics[width = .4\textwidth]{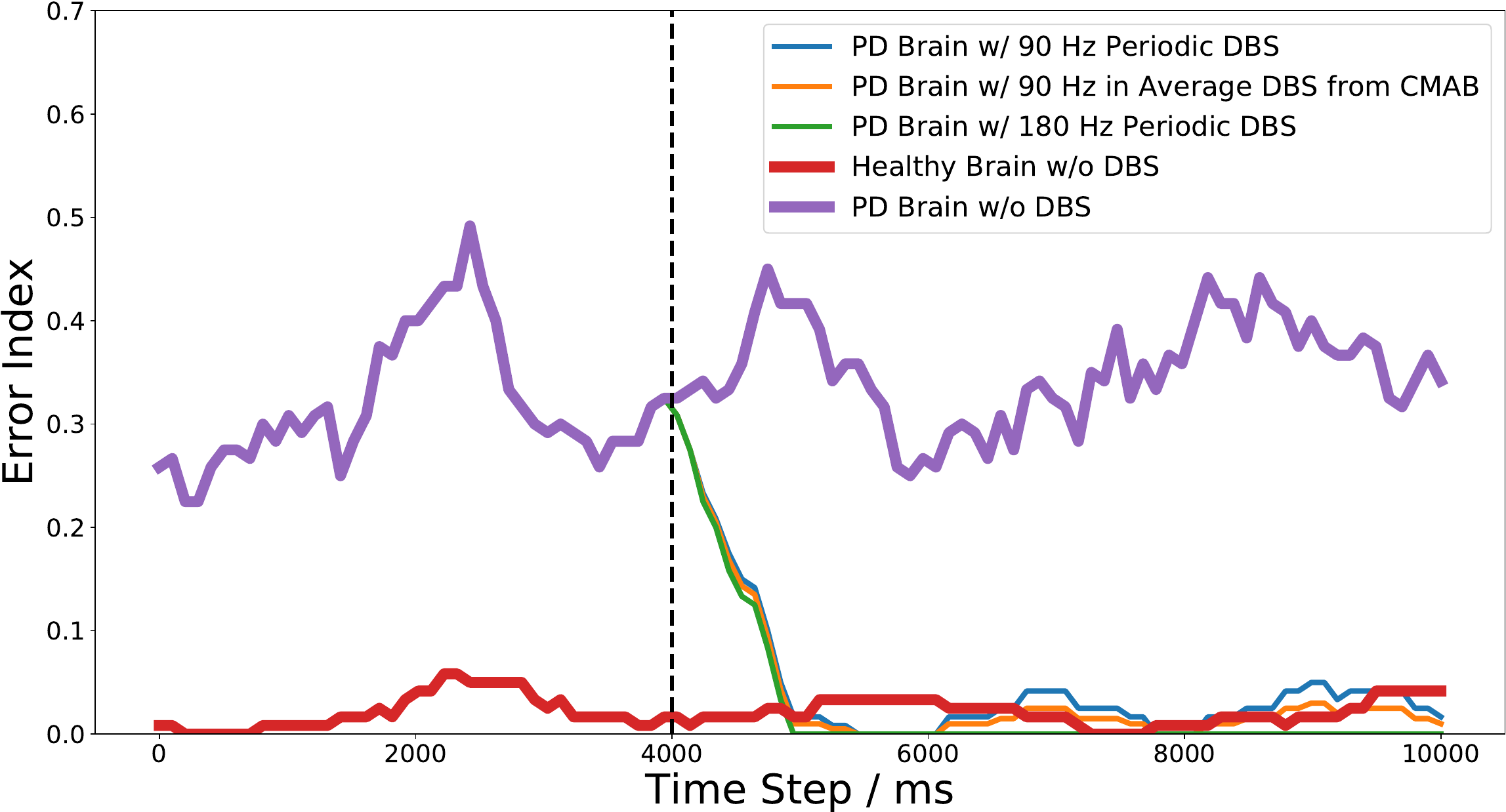}
    \caption{Error Index (EI) over time in model PD brains without and with various types of stimulation, as well as model
healthy brains. 
    }
    \label{fig:ei_visual}
\end{figure}

\begin{figure}[!t]
    \centering
    \includegraphics[width = .4\textwidth]{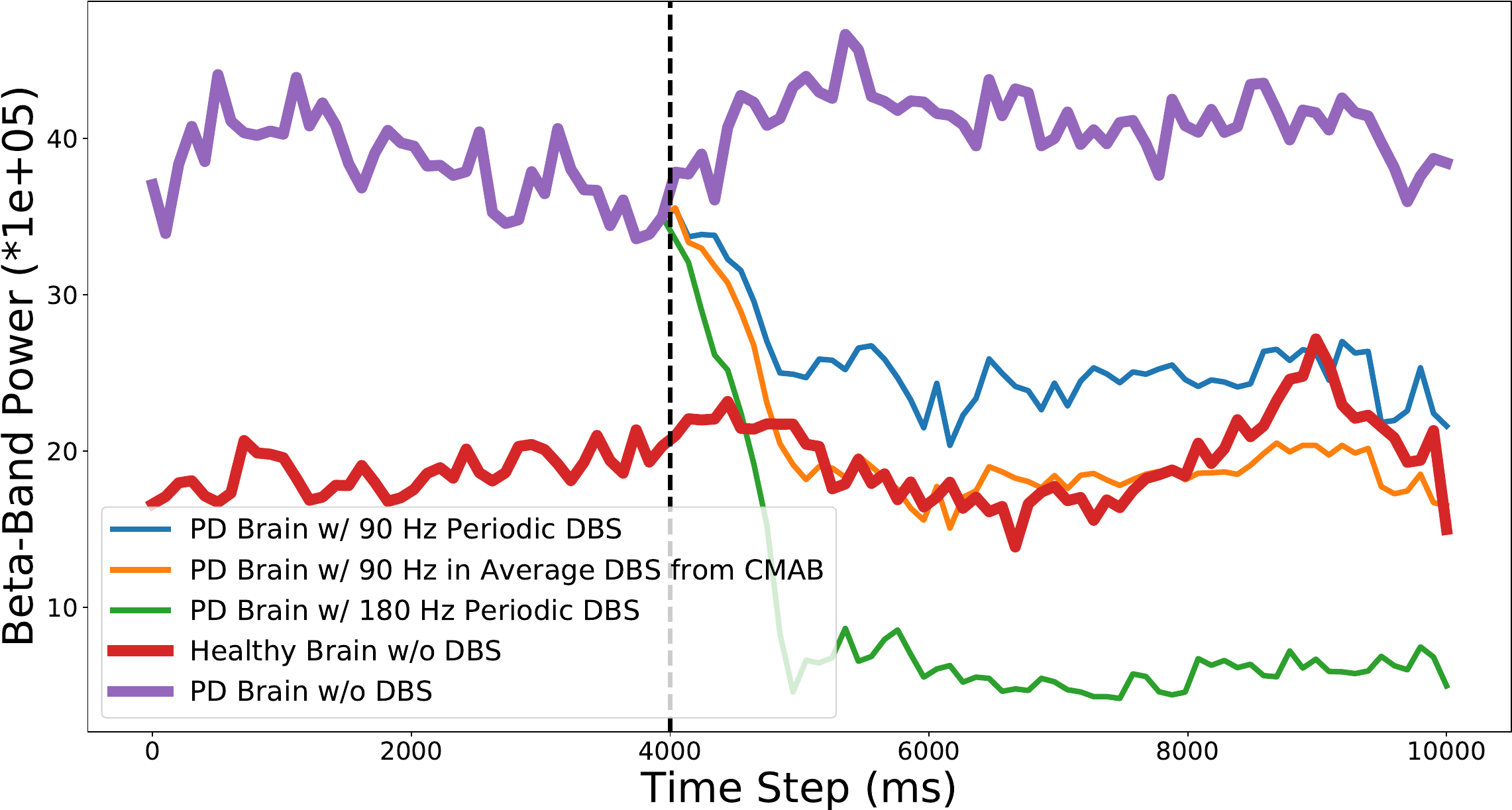}
    \caption{Beta power spectral density ($P_\beta$) over time in
model PD brains without and with various types of stimulation, as well as in healthy brains
    }
    \label{fig:beta_visual}
\end{figure}

\subsection{$\epsilon$-NeuralTS against Classical Controllers}
In addition to comparing with existing CMAB methods, we compared our $\epsilon$-NeuralTS ($\epsilon=0.8$) with the periodic controllers for which the stimulation pulses are equally spaced in terms of time steps. We firstly calculate the average stimulation frequency of $\epsilon$-NeuralTS ($\epsilon=0.8$) after convergence as about $k = 6$, \textit{i.e.,} $90 Hz$, which is smaller than the best setting of the vanilla NeuralTS in \Cref{table:coeff_freq}. In other words, less exploration can also help reduce the frequency of the stimulation, improving energy efficiency and reducing side effects. To have a fair comparison, we mainly compared our CMAB-based controller with periodic cDBS at $90 Hz$. 

In \Cref{fig:ei_visual} and \Cref{fig:beta_visual}, we evaluate the performance of our $\epsilon$-NeuralTS controller online after learning. Specifically, the whole evaluation period is divided by a dashed line, indicating that all DBS controllers will be turned on to output their corresponding execution after $4000$ rounds. Therefore, except for the healthy brain, all the other controllers start from the same oscillation with a higher EI and $P_\beta$. 

We observe that $90 Hz$ frequency is still high and effective enough for periodic cDBS, so periodic DBS at $90 Hz$ can easily reduce EI from the PD brain w/o DBS (purple curve) and even reaches a similar EI value as the periodic DBS at $180 Hz$ (\textit{i.e., } maximum value in our setting). Therefore, the difference between our method (orange) and periodic DBS at $90 Hz$ (blue) is not significant enough in \Cref{fig:ei_visual}. However, we do improve $P_\beta$ in \Cref{fig:beta_visual} compared with the periodic cDBS at $90 Hz$. Our method can successfully achieve a similar $P_\beta$ value as the healthy brain has, supporting that having varying stimulation frequency according to different observed context features can have better mitigation of PD symptoms even though the overall average frequency is the same.

\begin{figure}[!t]
    \centering
    \includegraphics[width = .45\textwidth]{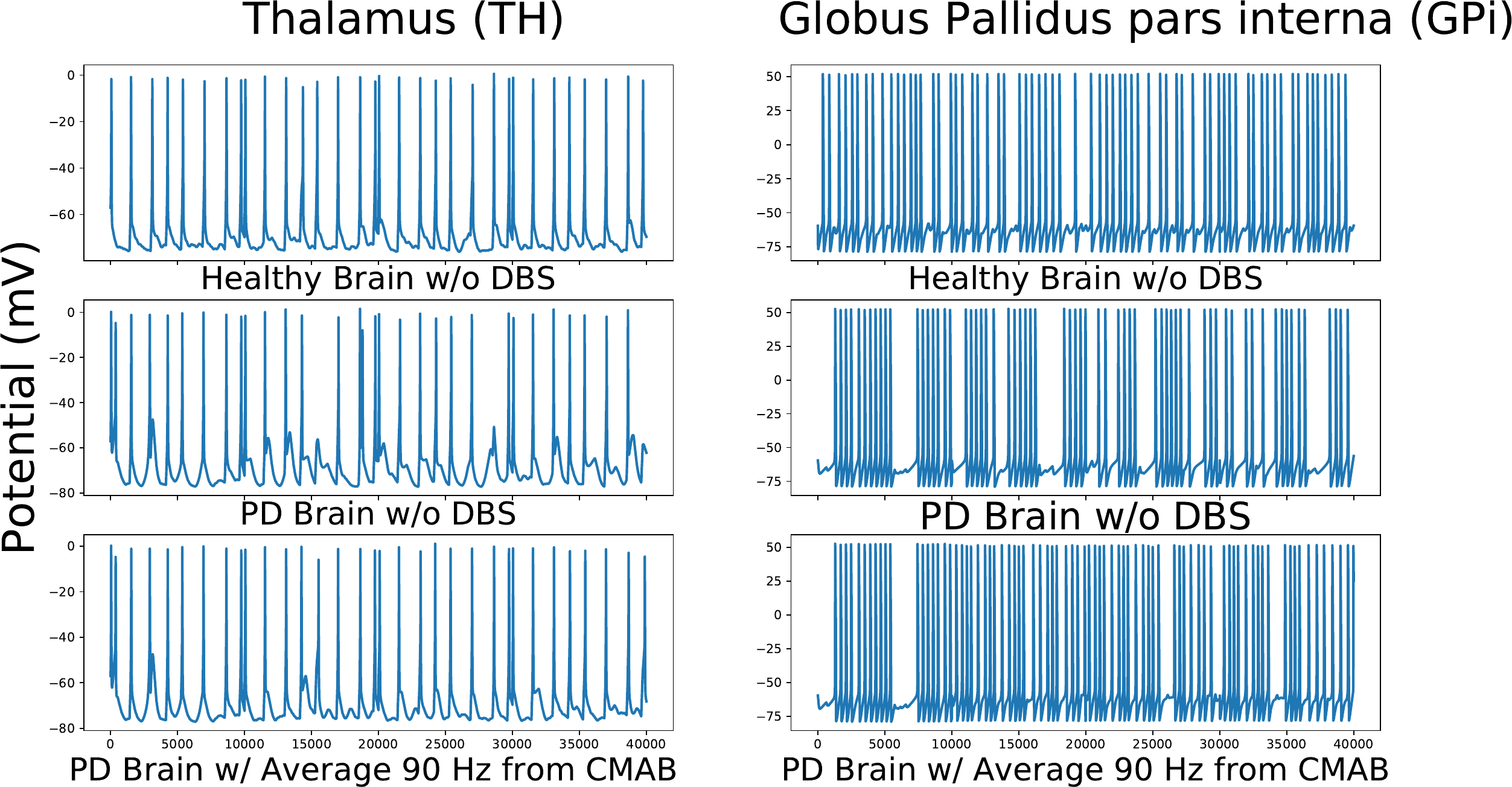}
    \caption{Activity of model neurons in TH
and GPi. The effects of pathophysiological patterns can be reduced using our $\epsilon$-NeuralTS with an average frequency of 90 Hz (bottom row). 
    }
    \label{fig:brain_potential}
\end{figure}

The TH neuron activations and the firing pattern in GPi in BGM will be different between healthy brain and a PD brain without DBS \cite{qitong2020}. In a healthy brain, the neuron activations of GPi and TH follow sporadic spiking at a stable firing rate. However, the brain affected by PD will result in pathological neuron activations within TH and GPi, which can be captured by reduced triggering potentials and clustered spiking, respectively. Therefore, in \Cref{fig:brain_potential}, we visualize the activity of model neurons in TH and GPi with a healthy brain, a PD brain without DBS, and a PD brain stimulated with our CMAB (\textit{i.e., } $\epsilon$-NeuralTS). 
We demonstrate that substantial pathophysiological patterns in the PD brain without DBS stimulation in the middle row of \Cref{fig:brain_potential} can be mitigated by our $\epsilon$-NeuralTS, which makes the pattern of the activity in both GPi and TH much similar to that of the healthy brain.

\begin{figure}[!t]
    \centering
    \subfigure[Reward (5 steps)]{
        \includegraphics[width =.23\textwidth]{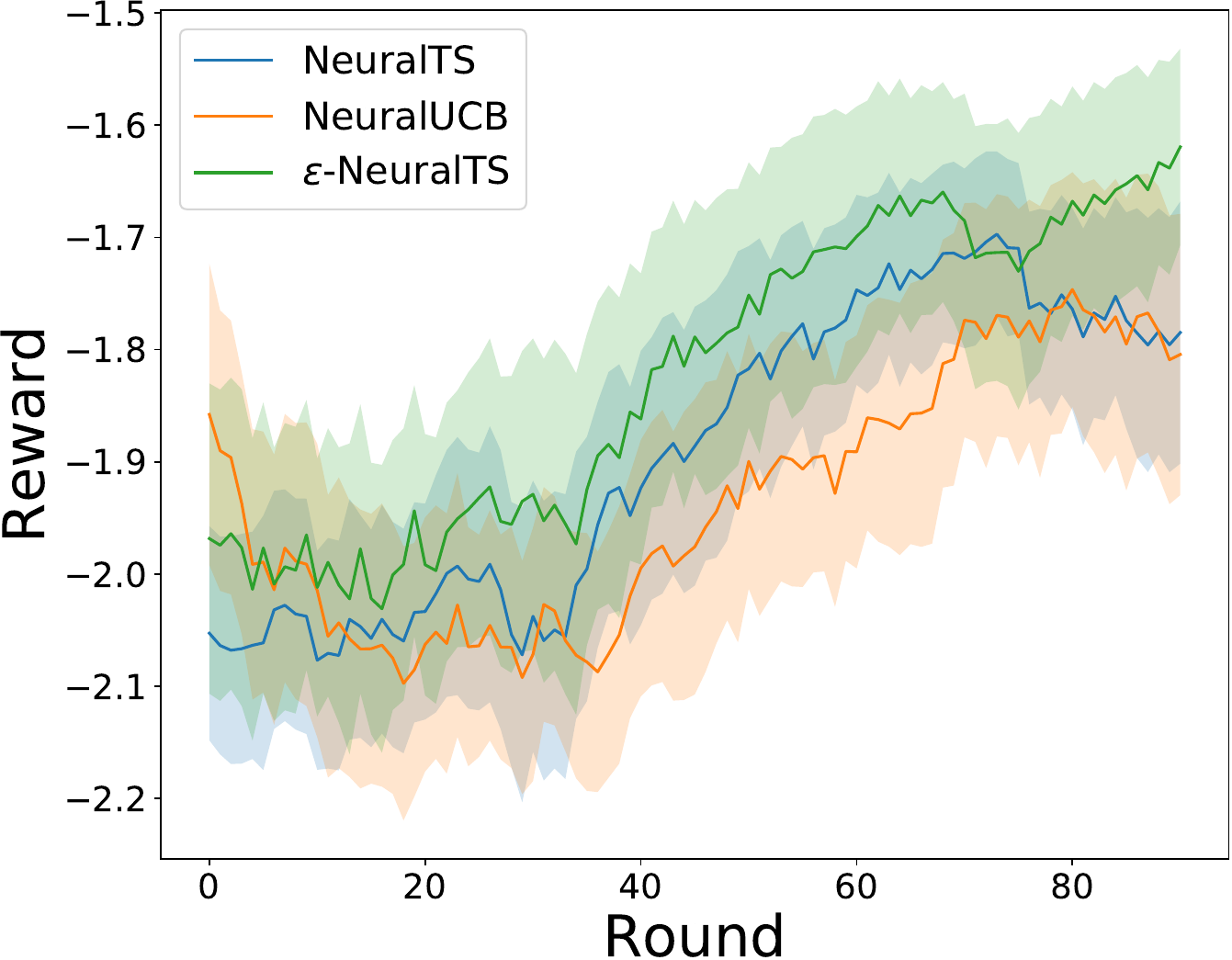}
    }
    \subfigure[Cumulative Regret (5 steps)]{
        \includegraphics[width =.22\textwidth]{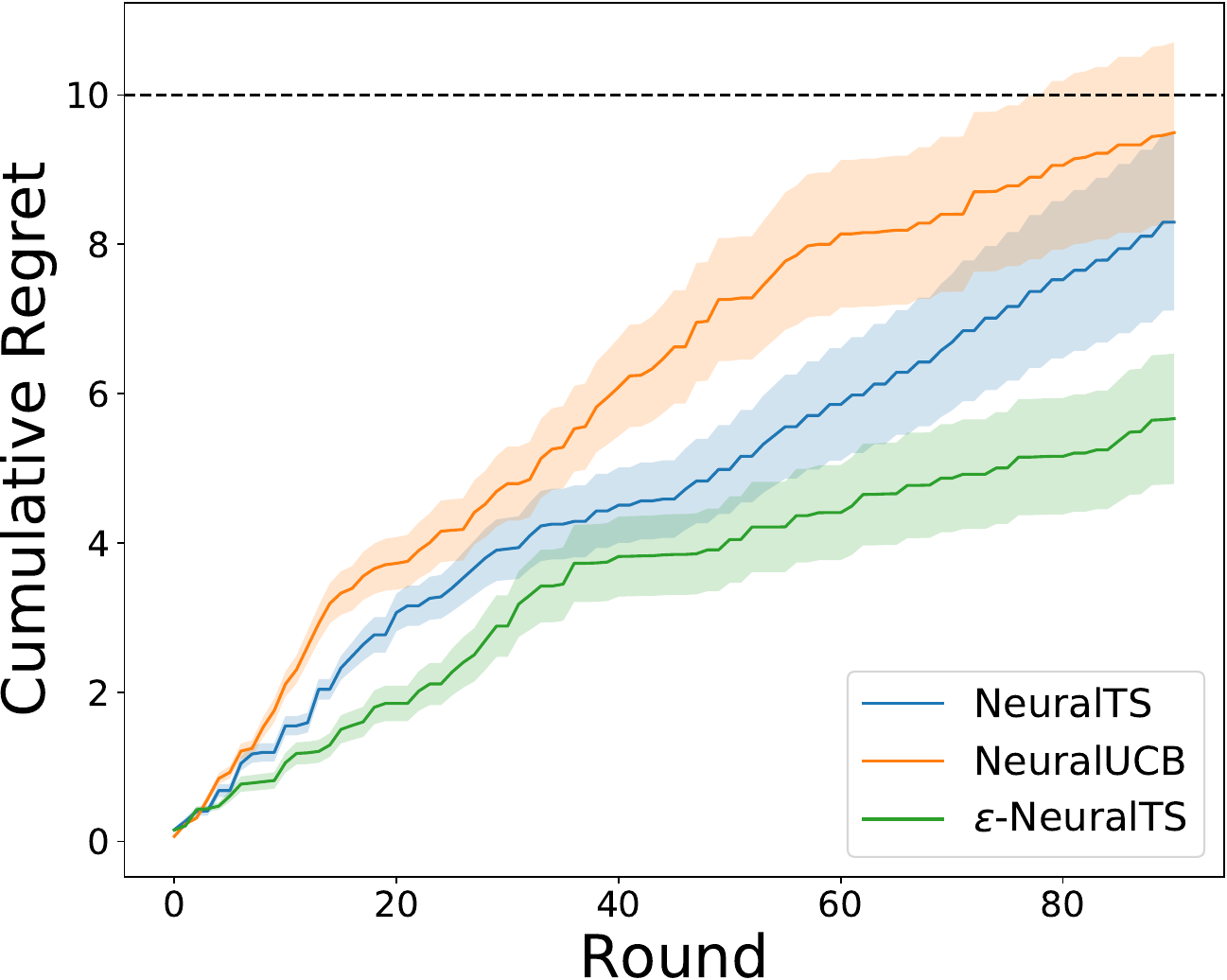}
     }
    \caption{  Comparison of $\epsilon$-NeuralTS with NeuralTS and NeuralUCB under 5 steps of delay. \textbf{Left:} rewards (higher is better) and \textbf{Right:} cumulative regret (lower is better) The total regret measures cumulative EI. Results are averaged over 10 runs with standard errors shown as shaded areas.}\label{fig:delay5}
\end{figure}

\begin{figure}[!t]
    \centering
       \subfigure[Reward (10 steps)]{
        \includegraphics[width =.23\textwidth]{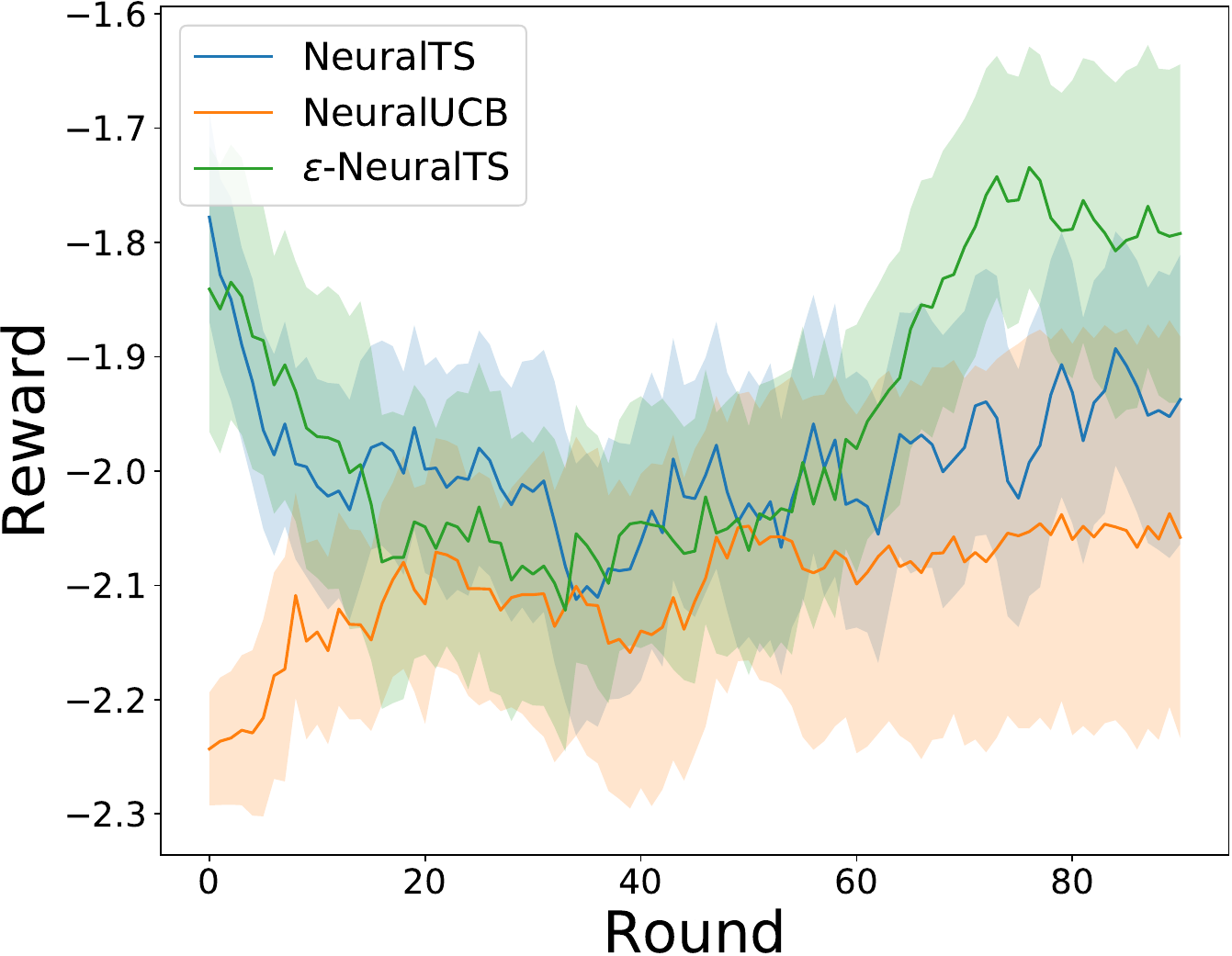}  
    }
    \subfigure[Cumulative Regret (10 step)]{
        \includegraphics[width =.22\textwidth]{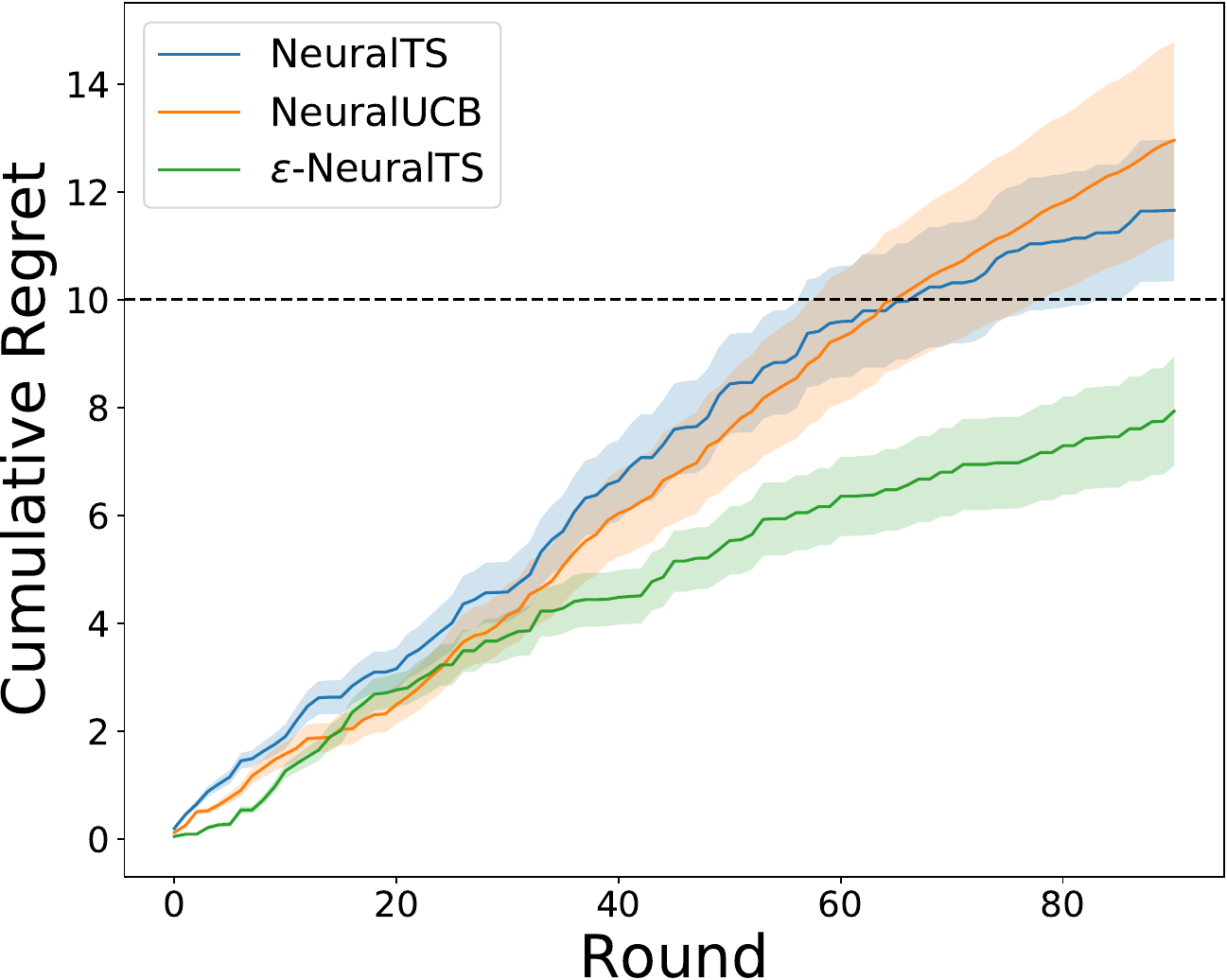}  
     }
    \caption{ Comparison of $\epsilon$-NeuralTS with NeuralTS and NeuralUCB under 10 steps of delay. \textbf{Left:} rewards (higher is better) and \textbf{Right:} cumulative regret (lower is better) The total regret measures cumulative EI. Results are averaged over 10 runs with standard errors shown as shaded areas.}\label{fig:delay10}
\end{figure}

\subsection{Robustness of $\epsilon$-NeuralTS (Reward Delay)}

Although data-driven approaches for decision-making (e.g., RL and CMAB) can demonstrate impressive performance in the training environment, they may not always provide a robust solution to internal conditions and external disturbances~\cite{rolah, robust_drone}. Specifically, this experiment is inspired by practical scenarios where the reward signals are delayed, due to various constraints when the algorithms are deployed in the real world~\cite{empirical_ts}. We study the robustness of the two most competitive CMAB from \Cref{sec:exp_CMAB} (i.e., $\epsilon$-NeuralTS and NeuralUCB) when the rewards are delayed. Particularly, the bandit learner will not receive the reward right after taking an action. Instead, the rewards will arrive in batches when the algorithm updates its~model.

In this evaluation, we only vary the batch size, which is the amount of the reward delay, considering $5$ and $10$ steps. Since Neural-TS is an instance of our $\epsilon$-NeuralTS, we also include its results for comparison. We report the reward and cumulative regret of three methods in \Cref{fig:delay5} and \Cref{fig:delay10}.

We firstly observe that NeuralTS outperforms NeuralUCB, which differs the standard setting without the reward delay. We notice that the gap between the vanilla NeuralTS and NeuralUCB in the standard task is small. Then according to~\cite{NeuralTS}, the core method for TS using randomized exploration encourages exploration between batches. Moreover, less exploration in $\epsilon$-NeuralTS also reduces delayed explorative information, leading to a better performance in both the task reward and cumulative regret. Specifically, the increase in the cumulative regret with 5 steps reward delay is not obvious for $\epsilon$-NeuralTS. When the amount of the reward delay is increased up to 10 steps, all of the learning rewards start to fluctuate, indicating that the reward delay does influence the learning process for all methods. Nevertheless, our $\epsilon$-NeuralTS still has a relatively smaller cumulative regret (below the dashed line) in \Cref{fig:delay10} compared to the other two approaches.

\section{Conclusion}\label{sec:conclusion}
Existing commercial DBS devices only support pre-defined periodic stimulation with fixed high frequency. To address this limitation, RL-based approaches have been proposed to search for a flexible and efficient stimulation frequency according to the status of the brain. Yet, in general, the use of RL requires a huge amount of training data. In addition, the required computational resources for learning RL policies in the real world also serves as an obstacle for deployment. Thus, in this work, we formulate the treatment of PD symptoms using DBS as a CMAB problem so that we only consider single-step decision-making, getting rid of the heavy computation required for RL training. We then propose a novel method using $\epsilon$-exploring strategy to reach more energy, sampling, and computationally efficient learning. Our method  outperforms the existing CMAB baselines and results in a lower $P_\beta$ compared to the periodic DBS with the same average stimulation frequency. The potential future direction is to investigate a better approximation of the mapping between $P_\beta$ and the oracle EI so that the bandit learner will tend to select the action with much lower frequency but with efficiency without access to EI. In addition, how to adapt our proposed method to real patient data will also be an avenue for future~work.



\bibliographystyle{IEEEtran}
\bibliography{ref}

\begin{thebibliography}{10}
\providecommand{\url}[1]{#1}
\csname url@samestyle\endcsname
\providecommand{\newblock}{\relax}
\providecommand{\bibinfo}[2]{#2}
\providecommand{\BIBentrySTDinterwordspacing}{\spaceskip=0pt\relax}
\providecommand{\BIBentryALTinterwordstretchfactor}{4}
\providecommand{\BIBentryALTinterwordspacing}{\spaceskip=\fontdimen2\font plus
\BIBentryALTinterwordstretchfactor\fontdimen3\font minus \fontdimen4\font\relax}
\providecommand{\BIBforeignlanguage}[2]{{%
\expandafter\ifx\csname l@#1\endcsname\relax
\typeout{** WARNING: IEEEtran.bst: No hyphenation pattern has been}%
\typeout{** loaded for the language `#1'. Using the pattern for}%
\typeout{** the default language instead.}%
\else
\language=\csname l@#1\endcsname
\fi
#2}}
\providecommand{\BIBdecl}{\relax}
\BIBdecl

\bibitem{Marras2018}
C.~Marras, J.~Beck, and et~al., ``Prevalence of parkinson’s disease across north america,'' in \emph{NPJ Parkinson’s disease}, 2018.

\bibitem{DBSPD2003}
A.~L. Benabid, ``Deep brain stimulation for parkinson’s disease,'' in \emph{Current opinion in neurobiology}, 2003, pp. 696--706.

\bibitem{randomDBS}
G.~Deuschl, C.~Schade-Brittinger, and et~al., ``A randomized trial of deep-brain stimulation for parkinson’s disease,'' in \emph{New England Journal of Medicine}, 2006, pp. 896--908.

\bibitem{pallidalDBS2010}
K.~A. Follett, F.~M. Weaver, and et~al., ``Pallidal versus subthalamic deep-brain stimulation for parkinson’s disease,'' in \emph{New England Journal of Medicine}, 2010, pp. 2077--2091.

\bibitem{DBSPD2012}
M.~S. Okun, ``Deep-brain stimulation for parkinson’s disease,'' in \emph{New England Journal of Medicine}, 2012, pp. 1529--1538.

\bibitem{Pineau2009}
J.~Pineau, A.~Guez, R.~Vincent, G.~Panuccio, and M.~Avoli, ``Treating epilepsy via adaptive neurostimulation: a reinforcement learning approach,'' in \emph{Int. Journal of Neural Systems}, 2009, pp. 227--240.

\bibitem{adbs2016}
M.~Beudel and P.~Brown, ``Adaptive deep brain stimulation in parkinson’s disease,'' in \emph{Parkinsonism \& related disorders}, 2016, pp. 123--126.

\bibitem{adbsChallenge2016}
M.~Arlotti, M.~Rosa, and et~al., ``The adaptive deep brain stimulation challenge,'' in \emph{Parkinsonism \& related disorders}, 2016, pp. 12--17.

\bibitem{adbsPD2016}
M.~Arlotti, L.~Rossi, and et~al., ``An external portable device for adaptive deep brain stimulation (adbs) clinical research in advanced parkinson’s disease,'' in \emph{Medical engineering \& physics}, 2016, pp. 498--505.

\bibitem{little2013}
L.~S, P.~A, and et~al., ``Adaptive deep brain stimulation in advanced parkinson disease,'' in \emph{Ann Neurol}, 2013, pp. 449--457.

\bibitem{little2016}
S.~Little, E.~Tripoliti, and et~al., ``Adaptive deep brain stimulation for parkinson’s disease demonstrates reduced speech side effects compared to conventional stimulation in the acute setting,'' in \emph{J Neurol Neurosurg Psychiatry}, 2016, pp. 1388--1389.

\bibitem{gao_iccps22}
Q.~Gao, S.~L. Schmidt, K.~Kamaravelu, D.~A. Turner, W.~M. Grill, and M.~Pajic, ``Offline policy evaluation for learning-based deep brain stimulation controllers,'' in \emph{13th ACM/IEEE International Conference on Cyber-Physical Systems (ICCPS)}, 2022, pp. 80--91.

\bibitem{gao_iccps23}
Q.~Gao, S.~L. Schmidt, A.~Chowdhury, G.~Feng, J.~J. Peters, K.~Genty, W.~M. Grill, D.~A. Turner, and M.~Pajic, ``Offline learning of closed-loop deep brain stimulation controllers for parkinson disease treatment,'' in \emph{ACM/IEEE 14th International Conference on Cyber-Physical Systems (ICCPS)}, 2023, pp. 44--55.

\bibitem{schmidt_brain23}
S.~L. Schmidt, A.~H. Chowdhury, K.~T. Mitchell, J.~J. Peters, Q.~Gao, H.-J. Lee, K.~Genty, S.-C. Chow, W.~M. Grill, M.~Pajic, and D.~A. Turner, ``{At home adaptive dual target deep brain stimulation in Parkinson's disease with proportional control},'' \emph{Brain}, vol. 147, no.~3, pp. 911--922, 12 2023.

\bibitem{habets2018}
J.~Habets, M.~Heijmans, and et~al., ``An update on adaptive deep brain stimulation in parkinson’s disease,'' in \emph{Movement Disorders}, 2018, pp. 1834--1843.

\bibitem{parisa2022}
P.~Sarikhani, H.-L. Hsu, and B.~Mahmoudi, ``Automated tuning of closedloop neuromodulation control systems using bayesian optimization,'' in \emph{2022 44th Annual International Conference of the IEEE Engineering in Medicine \& Biology Society (EMBC)}, 2022, pp. 1734--1737.

\bibitem{seizure2017}
V.~Nagaraj, A.~Lamperski, and T.~I. Netoff, ``Seizure control in a computational model using a reinforcement learning stimulation paradigm,'' in \emph{International J. of Neural Sys}, 2017.

\bibitem{qitong2020}
Q.~Gao, M.~Naumann, I.~Jovanov, V.~Lesi, K.~Kumaravelu, W.~Grill, and M.~Pajic, ``Model-based design of closed loop deep brain stimulation controller using reinforcement learning,'' in \emph{ACM/IEEE 11th Int. Conf. on Cyber-Physical Systems (ICCPS)}, 2020, pp. 108--118.

\bibitem{analy_bandit}
P.~Auer, N.~Cesa-Bianchi, and P.~Fischer, ``Finite-time analysis of the multiarmed bandit problem,'' in \emph{Machine Learning}, 2002, pp. 235--256.

\bibitem{hedging_rl}
L.~Cannelli, G.~Nuti, M.~Sala, and O.~Szehr, ``Hedging using reinforcement learning: Contextual k-armed bandit versus q-learning,'' \emph{The Journal of Finance and Data Science}, vol.~9, 2023.

\bibitem{commercial}
T.~Y and et~al., ``Towards adaptive deep brain stimulation: clinical and technical notes on a novel commercial device for chronic brain sensing,'' in \emph{Journal of Neural Engineering}, vol.~18, 2021.

\bibitem{beta_power}
K.~Kumaravelu, D.~T. Brocker, and W.~M. Grill, ``A biophysical model of the cortex-basal ganglia-thalamus network in the 6-ohda lesioned rat model of parkinson’s disease,'' in \emph{Journal of computational neuroscience}, 2016, pp. 207--229.

\bibitem{bgm2012}
R.~Q. So, A.~R. Kent, and W.~M. Grill, ``Relative contributions of local cell and passing fiber activation and silencing to changes in thalamic fidelity during deep brain stimulation and lesioning: a computational modeling study,'' in \emph{Journal of computational neuroscience}, vol.~32, 2012, pp. 499--519.

\bibitem{jovanov_iccps18}
I.~{Jovanov}, M.~{Naumann}, K.~{Kumaravelu}, W.~M. {Grill}, and M.~{Pajic}, ``Platform for model-based design and testing for deep brain stimulation,'' in \emph{ACM/IEEE 9th International Conference on Cyber-Physical Systems (ICCPS)}, April 2018, pp. 263--274.

\bibitem{thompson}
W.~R. Thompson, ``On the likelihood that one unknown probability exceeds another in view of the evidence of two samples,'' in \emph{Biometrika}, 1933, pp. 285--294.

\bibitem{error_index}
D.~T. Brocker and et~al., ``Optimized temporal pattern of brain stimulation designed by computational evolution,'' in \emph{Science Translational Medicine}, 2017.

\bibitem{bandit}
T.~Lattimore and C.~Szepesv´ari, ``Bandit algorithms,'' \emph{Cambridge University Press}, 2018.

\bibitem{LinTS}
S.~Agrawal and N.~Goyal, ``Thompson sampling for contextual bandits with linear payoffs,'' in \emph{International Conference on Machine Learning}, 2013, pp. 127--135.

\bibitem{foundationML}
S.~Bubeck and N.~Cesa-Bianchi, ``Regret analysis of stochastic and nonstochastic multi-armed bandit problems,'' in \emph{Foundations and Trends in Machine Learning}, 2011, pp. 1--122.

\bibitem{NIPS2011_e1d5be1c}
Y.~Abbasi-yadkori, D.~P\'{a}l, and C.~Szepesv\'{a}ri, ``Improved algorithms for linear stochastic bandits,'' in \emph{Advances in Neural Information Processing Systems}, J.~Shawe-Taylor, R.~Zemel, P.~Bartlett, F.~Pereira, and K.~Weinberger, Eds., vol.~24.\hskip 1em plus 0.5em minus 0.4em\relax Curran Associates, Inc., 2011.

\bibitem{LinUCB}
L.~Li, W.~Chu, J.~Langford, and R.~E. Schapire, ``A contextual-bandit approach to personalized news article recommendation,'' in \emph{{Proc. of the 19th Int. Conf. on World Wide Web}}, 2010, pp. 661--670.

\bibitem{NIPS2010_c2626d85}
S.~Filippi, O.~Cappe, A.~Garivier, and C.~Szepesv\'{a}ri, ``Parametric bandits: The generalized linear case,'' in \emph{Advances in Neural Information Processing Systems}, J.~Lafferty, C.~Williams, J.~Shawe-Taylor, R.~Zemel, and A.~Culotta, Eds., vol.~23.\hskip 1em plus 0.5em minus 0.4em\relax Curran Associates, Inc., 2010.

\bibitem{random2020bandit}
B.~Kveton, M.~Zaheer, C.~Szepesvari, L.~Li, M.~Ghavamzadeh, and C.~Boutilier, ``Randomized exploration in generalized linear bandits,'' in \emph{International Conference on Artificial Intelligence and Statistics}, 2020, pp. 2066--2076.

\bibitem{Li2017ProvablyOA}
L.~Li, Y.~Lu, and D.~Zhou, ``Provably optimal algorithms for generalized linear contextual bandits,'' in \emph{International Conference on Machine Learning}, 2017.

\bibitem{Ding2020AnEA}
Q.~Ding, C.-J. Hsieh, and J.~Sharpnack, ``An efficient algorithm for generalized linear bandit: Online stochastic gradient descent and thompson sampling,'' \emph{ArXiv}, vol. abs/2006.04012, 2020.

\bibitem{Riquelme2018}
C.~Riquelme, G.~Tucker, and J.~Snoek, ``Deep bayesian bandits showdown: An empirical comparison of bayesian deep networks for thompson sampling.'' in \emph{International Conference on Learning Representation (ICLR)}, 2018.

\bibitem{NeuralTS}
W.~Zhang, D.~Zhou, L.~Li, and Q.~Gu, ``Neural thompson sampling,'' in \emph{International Conference on Learning Representations}, 2021.

\bibitem{NeuralUCB}
D.~Zhou, L.~Li, and Q.~Gu, ``Neural contextual bandits with ucb-based exploration,'' in \emph{International Conference on Machine Learning}, 2020, pp. 11\,492--11\,502.

\bibitem{Xu2020NeuralCB}
P.~Xu, Z.~Wen, H.~Zhao, and Q.~Gu, ``Neural contextual bandits with deep representation and shallow exploration,'' in \emph{International Conference on Learning Representations}, 2022.

\bibitem{empirical_ts}
O.~Chapelle and L.~Li, ``An empirical evaluation of thompson sampling,'' in \emph{Advances in Neural Information Processing Systems}.\hskip 1em plus 0.5em minus 0.4em\relax Curran Associates, Inc., 2011, pp. 2249--2257.

\bibitem{SaferlHsu2022}
H.-L. Hsu, Q.~Huang, and S.~Ha, ``Improving safety in deep reinforcement learning using unsupervised action planning,'' in \emph{2022 IEEE International Conference on Robotics and Automation (ICRA)}, 2022, pp. 5567--5573.

\bibitem{neuroweaver}
P.~Sarikhani, H.-L. Hsu, O.~Kara, J.~K. Kim, H.~Esmaeilzadeh, and B.~Mahmoudi, ``Neuroweaver: a platform for designing intelligent closed-loop neuromodulation systems,'' \emph{Brain Stimulation}, vol.~14, no.~6, p. 1661, 2021.

\bibitem{eps_TS}
T.~Jin, X.~Yang, X.~Xiao, and P.~Xu, ``Thompson sampling with less exploration is fast and optimal,'' in \emph{International Conference on Machine Learning}, 2023.

\bibitem{Jin2024MATS}
T.~Jin, H.-L. Hsu, W.~Chang, and P.~Xu, ``Finite-time frequentist regret bounds of multi-agent thompson sampling on sparse hypergraphs,'' in \emph{Annual AAAI Conference on Artificial Intelligence (AAAI)}, 2024.

\bibitem{adaptiveRL2008}
A.~Guez, R.~D. Vincent, M.~Avoli, and J.~Pineau, ``Adaptive treatment of epilepsy via batch-mode reinforcement learning,'' in \emph{AAAI}, 2008, pp. 1671--1678.

\bibitem{UCBGLM}
L.~Li, Y.~Lu, and D.~Zhou., ``Provably optimal algorithms for generalized linear contextual bandits,'' in \emph{International Conference on Machine Learning}, 2017, pp. 2071--2080.

\bibitem{NeuralGreedy}
C.~Riquelme, G.~Tucker, and J.~Snoek., ``Deep bayesian bandits showdown: An empirical comparison of bayesian deep networks for thompson sampling,'' in \emph{International Conference on Learning Representations}, 2018.

\bibitem{rolah}
J.~Dong, H.-L. Hsu, Q.~Gao, V.~Tarokh, and M.~Pajic, ``Robust reinforcement learning through efficient adversarial herding,'' \emph{https://arxiv.org/abs/2306.07408}, 2023.

\bibitem{robust_drone}
H.-L. Hsu, H.~Meng, S.~Luo, J.~Dong, V.~Tarokh, and M.~Pajic, ``Reforma: Robust reinforcement learning via adaptive adversary for drones flying under disturbances,'' in \emph{2024 IEEE International Conference on Robotics and Automation (ICRA)}, 2024.

\end{thebibliography}

\newpage
\onecolumn

\section*{Appendix}
To improve readability of the paper, we provide a summary of the employed notation in \Cref{table:notation}.

\begin{table*}[!t]
\begin{center}
\caption{The notation used in the paper.}
\label{table:notation}
\begin{tabular}{cp{12cm}}
\toprule
Symbol & Description \\[0.5ex] 
\midrule
$n$ & number of neurons in each sub-region $q$ in the brain\\
$\nu_j^q$ & $j^{th}$ neuron's electrical potential with the corresponding sub-region $q \in \{STN, GPe, GPi, TH\}$\\
$\bm{v}^q(t)$ & vector of electrical potential with the corresponding sub-region $q \in \{STN, GPe, GPi, TH\}$ at time $t$\\
EI & Error Index: portion of erroneous TH neuron activations in response to SMC inputs\\
$P_\beta$ & Beta-band Power Spectral Density\\

$T$ & maximum number of rounds defined as a horizon in multi-armed bandit problem\\
$K$ & number of arms\\
$A$ & action set\\
$a_t$ & action in time $t$ in multi-armed bandit problem\\
$u_t$ & mapped action from $a_t$ to computational BGM at time $t$\\
$s_t$ & context feature in time $t$\\
$x_t$ & context vector transformed from $s_t$ in time $t$\\
$\pi$ & policy\\
$R_{t,a_t}$ & reward in time $t$ corresponding to $a_t$\\
$D$ & history storing a sequence of tuple ($a_t$, $R_t$)\\ 
$T_w$ & window of size\\
$r_(T)$ & cumulative regret up to time $T$\\
$\nu >0$ & exploration variance\\ 
$\lambda$ & regularization parameter\\ 
$\epsilon$ & exploration probability\\

\bottomrule
\end{tabular}
\end{center}
\end{table*}

\end{document}